\documentclass{article}

\usepackage{microtype}
\usepackage{graphicx}
\usepackage{subcaption}
\usepackage{booktabs} 

\usepackage{hyperref}

\usepackage[accepted]{icml2026}

\usepackage{amsmath}
\usepackage{amssymb}
\usepackage{mathtools}
\usepackage{amsthm}
\usepackage{xspace}

\usepackage{pifont}
\newcommand{\cmark}{\ding{51}}
\newcommand{\xmark}{\ding{55}}

\usepackage[capitalize,noabbrev]{cleveref}

\theoremstyle{plain}

\theoremstyle{definition}

\theoremstyle{remark}

\newcommand{\model}{OmniMoE\xspace}
\newcommand{\method}{OmniMoE\xspace}

\usepackage[textsize=tiny]{todonotes}

\icmltitlerunning{\method}

\begin{document}

\twocolumn[
  \icmltitle{OmniMoE: An Efficient MoE by Orchestrating Atomic Experts at Scale}

  \icmlsetsymbol{equal}{*}

  \begin{icmlauthorlist}
    \icmlauthor{Jingze Shi}{HKUST(GZ)}
    \icmlauthor{Zhangyang Peng}{HKUST(GZ)}
    \icmlauthor{Yizhang Zhu}{HKUST(GZ)}
    \icmlauthor{Yifan Wu}{HKUST(GZ)}
    \icmlauthor{Guang Liu}{BAAI}
    \icmlauthor{Yuyu Luo}{HKUST(GZ)}
  \end{icmlauthorlist}

  \icmlaffiliation{HKUST(GZ)}{The Hong Kong University of Science and Technology (Guangzhou)}
  \icmlaffiliation{BAAI}{Beijing Academy of Artificial Intelligence}

  \icmlcorrespondingauthor{Yuyu Luo}{yuyuluo@hkust-gz.edu.cn}

  \icmlkeywords{Machine Learning, ICML}

  \vskip 0.3in
]



\printAffiliationsAndNotice{}  

\begin{abstract}
Mixture-of-Experts (MoE) architectures are evolving towards finer granularity to improve parameter efficiency. However, existing MoE designs face an inherent trade-off between the granularity of expert specialization and hardware execution efficiency.
We propose \textbf{OmniMoE}, a system–algorithm co-designed framework that pushes expert granularity to its logical extreme. OmniMoE introduces vector-level \textbf{Atomic Experts}, enabling scalable routing and execution within a single MoE layer, while retaining a shared dense MLP branch for general-purpose processing.
While this \emph{atomic} design maximizes capacity, it poses severe challenges for routing complexity and memory access. To address these, \method adopts a system-algorithm co-design: (i)~a \textbf{Cartesian Product Router} that decomposes the massive index space to reduce routing complexity from $O(N)$ to $O(\sqrt{N})$; and (ii)~\textbf{Expert-Centric Scheduling} that inverts the execution order to turn scattered, memory-bound lookups into efficient dense matrix operations. Validated on seven benchmarks, \method (with 1.7B active parameters) achieves 50.9\% zero-shot accuracy across seven benchmarks, outperforming coarse-grained and fine-grained baselines. Crucially, \model reduces inference latency from 73ms to 6.7ms (a 10.9$\times$ speedup) compared to PEER, demonstrating that massive-scale fine-grained MoE can be fast and accurate. 
Our code is open-sourced at \url{https://github.com/HKUSTDial/omni-moe}.
\end{abstract}

\section{Introduction}
\label{sec:introduction}

\begin{figure}[t!]
    \centering
    \includegraphics[width=0.94\linewidth]{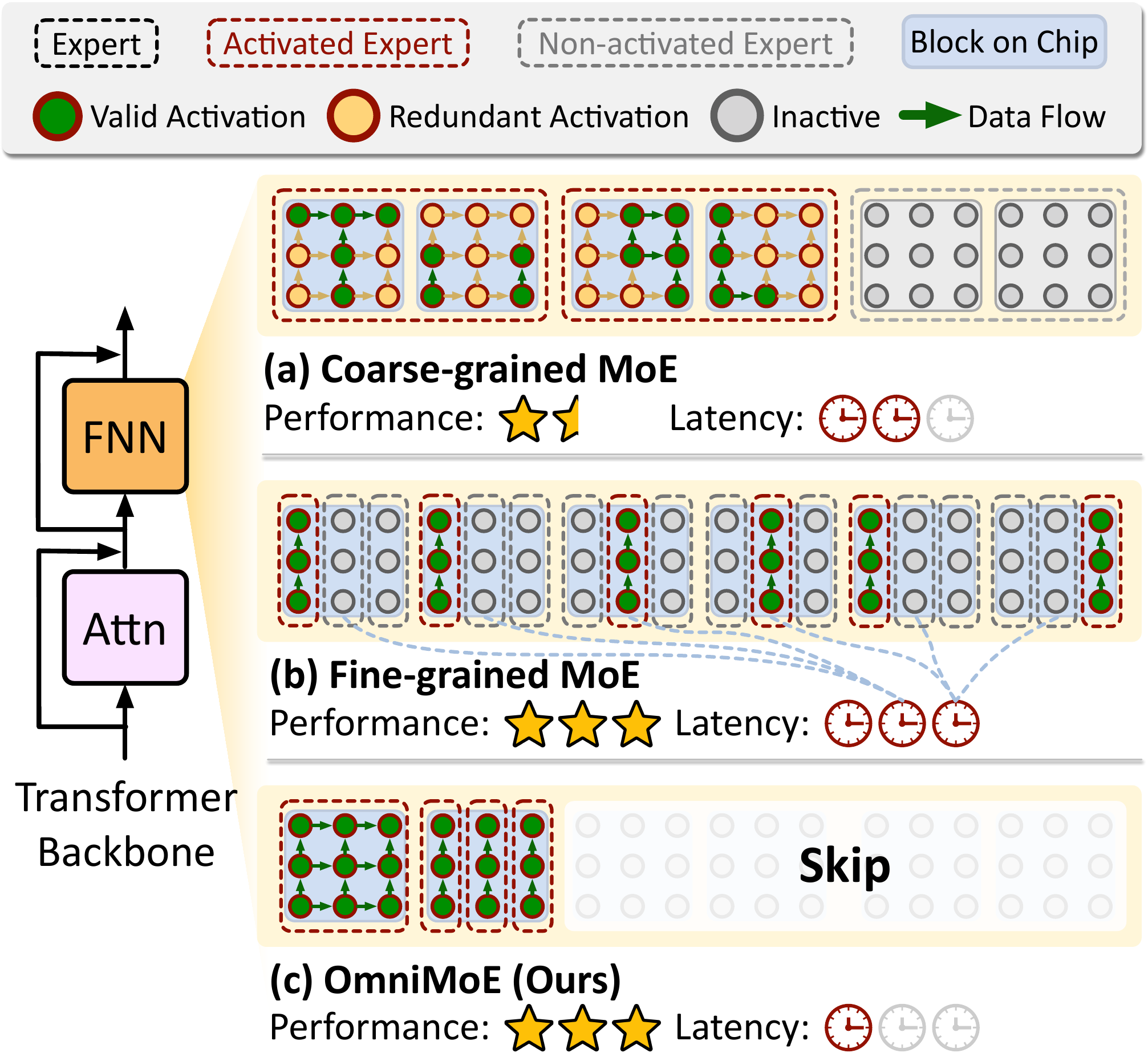}
    \caption{
        \textbf{Activation Patterns and System Optimization.}
        (a) Coarse-grained MoE activates large experts, inevitably involving redundant parameters and wasting computation. (b) Fine-grained MoE improves parameter efficiency, but suffers from bandwidth bottlenecks due to scattered, fragmented memory accesses. (c) Our \model employs a universally activated shared dense MLP, and uses expert-centric scheduling to reorganize fine-grained expert fetches into contiguous, coalesced memory accesses, achieving both high parameter efficiency and hardware-efficient execution.
    }
    \label{fig:teaser}
    \vspace{-2em}
\end{figure}

Mixture-of-Experts (MoEs) has emerged as a key approach to mitigating scaling bottlenecks by partially decoupling model capacity from per-token computation~\cite{fedus22switch}.
By activating only a subset of experts for each token, MoEs allow for massive parameter scaling while maintaining manageable inference budgets. 
A central design choice in MoE is the \emph{granularity} of experts, which largely determines both routing precision and system efficiency. Broadly, existing designs fall into two categories: coarse-grained MoEs and fine-grained MoEs.

\noindent\textbf{Coarse-Grained MoEs.}
Coarse-grained architectures represent the dominant paradigm in contemporary large-scale language models. Representative systems~\cite{du2022glam, jiang2024mixtral, zoph2022st} such as DeepSeek-V3~\cite{deepseek2025deepseekv3} (256 experts) and KIMI-K2~\cite{kimi2025k2} (384 experts) instantiate each expert as a complete dense FFN, benefiting from hardware-efficient dense matmuls (via Tensor Cores), contiguous VRAM access, and 
shared-expert general knowledge and training stability
~\cite{dai2024deepseekmoe,nguyen2025ondeepseekmoe, deepseek2025deepseekv3,kimi2025k2}.
Despite the success, coarse-grained MoEs inherently suffer from \textit{imprecise activation}~\cite{szatkowski2024d2dmoe} and \textit{low flexibility}. Specifically, activating large expert blocks incurs computation on parameters irrelevant to specific tokens (orange nodes in Figure~\ref{fig:teaser} (a)), leading to computational waste~\cite{cheng2025mone, li2023lazyneuron, szatkowski2024d2dmoe}.
Moreover, their rigid size hinders adaptation to limited hardware: 
coarse granularity restricts scaling flexibility, forcing steep, discrete memory increments when adjusting expert counts.

\noindent\textbf{Fine-grained MoEs.} Fine-grained architectures seek to maximize expressivity by utilizing millions of lightweight experts (e.g., embeddings).
MoE scaling-law analyses~\cite{krajewski2024scalinglaw, clark2022scalinglaw} suggest that, under a fixed training-token budget, performance improves with the total number of activated experts.
This motivates \emph{fine-grained} MoEs~\cite{he2024peer, nogueira2024mowe} that use extra lightweight experts. For example, PEER~\cite{he2024peer} scales to millions of experts by adopting a Product Key Memory (PKM~\cite{lample2019pkm}) style design, enabling precise routing and fine-grained control over both model capacity and activated parameters through smooth scaling.

However, scaling fine-grained experts to massive magnitudes introduces three system challenges.
(i) \textbf{Limited expressivity:} existing designs (e.g., PEER~\cite{he2024peer}) reduce experts to static parameter vectors. This restricts the expert computation to linear vector aggregation, stripping away the token-dependent nonlinear transformations (e.g., MLP projections) essential for modeling complex linguistic dependencies.
(ii) \textbf{Routing overhead:} scaling to a large expert pool increases routing cost and load imbalance, resulting in skewed expert utilization at scale.
(iii) \textbf{Hardware inefficiency:} scattered activations trigger random memory I/O, shifting execution from compute-bound to memory-bound and degrading GPU utilization.
As illustrated in Figure~\ref{fig:teaser}(b), while fine-grained experts ensure precise activation, the active parameters are inherently scattered across memory, which triggers frequent, non-contiguous memory accesses, inevitably shifting the execution bottleneck from computation to memory bandwidth.

While coarse-grained MoEs benefit from hardware-friendly architecture, the fine-grained ones leverage high activation efficiency and flexibility.
This raises a key question: \textbf{\textit{Is it possible to reconcile the parameter efficiency of fine-grained models with the hardware efficiency of coarse-grained architectures?}}
Realizing this synergy is non-trivial. It requires a holistic orchestration that simultaneously enhances the expressivity of fine-grained experts, minimizing routing overhead in large expert spaces, and reshaping irregular sparse accesses into hardware-efficient execution.

\begin{figure*}[!t]
    \centering
    \includegraphics[width=0.88\linewidth]{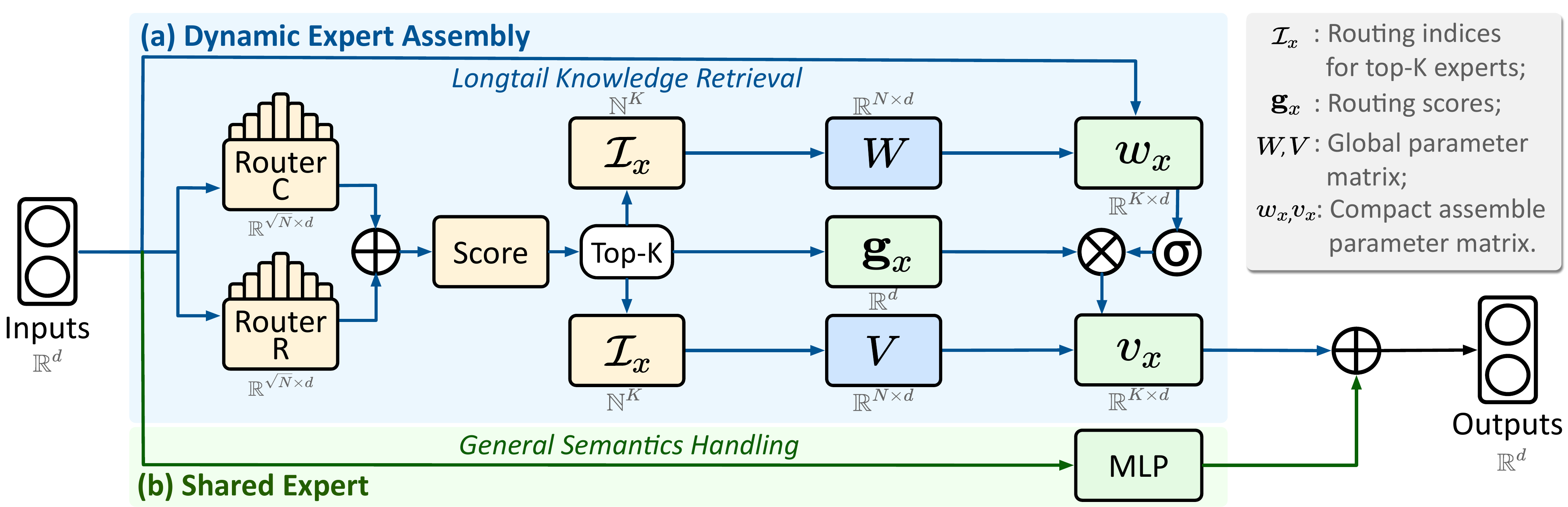}
    \vspace{-.5em}
    \caption{
      \textbf{Overview of the \method Architecture.}
        The framework operates via two parallel pathways to balance efficiency and expressivity.
        \textbf{(a) Dynamic Expert Assembly (Top):} For \textit{Longtail Knowledge Retrieval} objective, we employ a \textbf{Cartesian Product Router} (decomposed into Row/Column routers) to efficiently compute routing scores $\mathbf{g}_x$ and identify the top-$K$ expert indices $\mathcal{I}_x$.
        Then the system dynamically retrieves specific parameter slices from the global matrices $W, V$ to assemble compact, token-dependent parameter blocks $w_x, v_x$ for the final gated projection.
        \textbf{(b) Shared Expert (Bottom):} A dense MLP which is always active to handling \textit{General Semantics}.
        The final output is obtained by aggregating the outputs from the sparse, routed branch and the shared dense branch.  
    }
    \vspace{-1.0em}
    \label{fig:flash_moe_flow}
\end{figure*}

\noindent\textbf{Our Methodology and Contributions.} To address the aforementioned challenges, we propose \textbf{\method}, a system-algorithm co-designed MoE framework that synergizes the precise parameter activation of fine-grained experts with the hardware efficiency of coarse-grained designs. The core architectural innovation lies in a hybrid parallel design that combines a shared dense MLP for capturing general semantic knowledge, with a massive pool of routed fine-grained experts that specialize in long-tail knowledge retrieval. To orchestrate the activation of these fine-grained experts at scale, we introduce three tightly integrated contributions that jointly resolve the key bottlenecks.

First, to maximize model capacity and routing precision, we push expert granularity to its logical extreme by introducing the \textbf{Atomic Expert}, a minimal routable unit parameterized by a pair of vectors, and propose a corresponding \textbf{Dynamic Expert Assembly (DEA)} mechanism to organize and compose massive experts. 
This formulation enables scaling to a massive expert pool and supports highly specialized, token-specific, \emph{high-expressivity} parameter compositions.

However, orchestrating large-scale atomic experts poses an unprecedented routing challenge: standard approaches would incur prohibitive \emph{routing overhead}. To address this, we introduce the \textbf{Cartesian Product Router}. It decomposes the massive, 1D expert index space into a two-dimensional grid. By factorizing the routing computation into two independent, low-dimensional projections, 
it reduces the routing cost from linear in $N$ to proportional to $\sqrt{N}$, making large-scale expert routing practical and efficient.

With routing no longer the bottleneck, the key challenge in our orchestration shifts to \emph{hardware inefficiency}: fine-grained routing induces highly scattered atomic expert accesses, leading to poor locality and low GPU efficiency.
To overcome this final obstacle, we developed \textbf{Expert-Centric Scheduling}. This systemic contribution inverts the execution paradigm from token-centric to expert-centric. By reordering computations, it groups requests targeting the same experts, thereby converting scattered memory lookups into contiguous, reusable reads and enabling the use of high-throughput Grouped GEMM operations.

In summary, OmniMoE is a system-algorithm co-designed framework that orchestrates fine-grained expert activation, from atomic expert formulation to efficient routing and hardware-aware scheduling, achieving both high model expressivity and hardware efficiency at scale.
As illustrated in Figure~\ref{fig:teaser}, our heterogeneous architecture eliminates redundant computation inherent to coarse-grained MoEs, while our scheduling strategy resolves bandwidth bottlenecks caused by scattered, non-contiguous memory accesses under fine-grained routing. Extensive experiments demonstrate that \method achieves superior performance with \textbf{10.9$\times$ speedup} compared to state-of-the-art baselines. Our code is open-sourced at \url{https://github.com/HKUSTDial/omni-moe}.

\section{Methodology}
\label{sec:methodology}

We first formalize the general MoE architecture. A standard MoE layer comprises a pool of $N$ experts $\mathbb{E} = \{E_1, \dots, E_N\}$ and a routing function $\mathcal{G}(\cdot):\mathbb{R}^{d}\rightarrow \mathbb{R}^{N}$. Several MoE variants~\cite{dai2024deepseekmoe, deepseek2025deepseekv3, qwen32025} incorporate a shared dense MLP that remains universally activated for all inputs. For each input token with representation $x \in \mathbb{R}^d$, where d is the dimension of hidden states, the router computes routing scores and selects a subset of $K$ experts identified by indices $\mathcal{I}_x$.
\begin{equation}
    \label{eq:indicies_matrix_def}
    \mathcal{I}_x = (I_0,\ldots,I_{K-1}) = \operatorname{TopK}(\mathcal{G}(x), K)
\end{equation}    
where $I_i$ is the index of the $i$-th activated expert. And the routing weight $g_i$ for each expert $E_{I_i}$ could be calculated by 
\begin{equation}
\label{eq:gating-score}
    g_{i} = \operatorname{Softmax}(\mathcal{G}(x)[\mathcal{I}_x])_{i}, i\in[0,K)
\end{equation}
Finally, the layer output is the weighted sum of these activated experts:
\begin{equation}
y = \sum_{i \in [0,K)} g_{i} \cdot E_{I_i}(x) + \text{MLP}(x)
\label{eq:general_moe}
\end{equation}

\noindent\textbf{\method Overview.}
Figure~\ref{fig:flash_moe_flow} illustrates the overall architecture of \method.
Our design follows the standard MoE formulation in Eq.~\ref{eq:general_moe}. 
Furthermore, \method instantiates the \emph{activated} experts as \textbf{Atomic Experts} (Section~\ref{subsec:atomic-experts}) and uses our Dynamic Expert Assembly (DEA) mechanism to retrieve and assemble token-conditioned parameters on the fly, enabling the routed branch to operate at a much finer granularity than conventional FFN experts.
In parallel, we retain a dense $\text{MLP}$ as a \emph{shared expert} to provide general semantic reasoning and stable capacity that is independent of routing. 
The final representation is obtained by summing the shared dense branch and the routed fine-grained branch. 
The remainder of this section introduces (i) how we parameterize and store atomic experts efficiently (Section~\ref{subsec:atomic-experts}), (ii) how we route over massive expert spaces (Section~\ref{subsec:cartesian_product_router}), and (iii) how we schedule the resulting sparse computations to maximize hardware efficiency (Section~\ref{subsec:expert_centric_scheduling}).

\subsection{Atomic Experts and Dynamic Expert Assembly}
\label{subsec:atomic-experts}

In this section, we introduce the core components of our fine-grained expert design. We first define the \textbf{atomic expert} as the minimal routable computational unit. We then present \textbf{Dynamic Expert Assembly (DEA)}, our proposed mechanism for logically organizing these atomic units. Specifically, DEA enables the model to dynamically retrieve a sparse set of atomic experts from a global pool and compose them into a token-conditioned \emph{assembled expert}.

An \textbf{atomic expert} $E_i$ is defined as a minimal, lightweight computational unit parameterized by an input vector $w^{in}_i \in \mathbb{R}^{d}$ and an output vector $w^{out}_i \in \mathbb{R}^{d}$. Given a token representation $x \in \mathbb{R}^{d}$, its computation is:
\begin{equation}
    E_i(x) = \sigma(x {w^{in}_i}^\top) w^{out}_i
    \label{eq:atomic_expert}
\end{equation}
where $\sigma(\cdot)$ is a non-linear activation. Throughout \method, we instantiate $\sigma(\cdot)$ with \textsc{SwiGLU}~\cite{noam2020swiglu}. 
While a single atomic expert exhibits limited expressivity, the strength of our approach arises from the dynamic composition of these experts.

The \textbf{Dynamic Expert Assembly (DEA)} mechanism governs this composition process. For each input token, DEA consists of two steps: (i) \textbf{Retrieval}, where it selects a sparse subset of the most relevant atomic experts from massive global experts, and (ii) \textbf{Assembly}, which composes the retrieved parameters into a computational block.

To make this DEA computationally feasible at scale, the parameters of all $N$ atomic experts are not stored individually. Instead, they are consolidated into two global parameter matrices, $W, V \in \mathbb{R}^{N \times d}$, which serve as a centralized parameter repository for efficient retrieval and composition.
\begin{equation}
    \label{eq:global_matrices}
    W = [w^{in}_0, \dots, w^{in}_{N-1}]^\top, \quad V = [w^{out}_0, \dots, w^{out}_{N-1}]^\top
\end{equation}
For a given input token $x$, the routing mechanism identifies $\mathcal{I}_x \subset \{0, ..., N-1\}$, the indices of the top-$K$ most relevant atomic experts, similar to Eq.~\ref{eq:indicies_matrix_def}. The \textbf{Retrieval} step of DEA is implemented by gathering the rows indexed by $\mathcal{I}_x$ from the global parameter matrices, yielding compact, token-local parameter blocks:
\begin{equation}
    w_{x} = W[\mathcal{I}_{x}] \in \mathbb{R}^{K \times d}, \quad v_{x} = V[\mathcal{I}_{x}] \in \mathbb{R}^{K \times d}
    \label{eq:retrieval}
\end{equation}
Simultaneously, the associated routing scores are collected into a vector $\mathbf{g}_x = [g_{I_0}, \ldots, g_{I_{K-1}}] \in \mathbb{R}^K$. The \textbf{Assembly} step is then performed by composing these retrieved parameters and their corresponding weights into a single, fused computation:
\begin{equation}
    y = (\mathbf{g}_x \odot \sigma(x w_{x}^\top)) v_{x} + \text{MLP}(x).
    \label{eq:assembly}
\end{equation}
This formulation demonstrates how DEA effectively constructs a unique, powerful assembled expert for each token by composing simple, reusable atomic experts. This approach ensures that every retrieved parameter is computationally active for the target token, achieving extreme parameter efficiency while maintaining high expressivity.

\subsection{Cartesian Product Router}
\label{subsec:cartesian_product_router}

The primary role of the router is to efficiently select a sparse set of atomic expert indices $\mathcal{I}$ for the DEA mechanism.
However, scaling to massive expert pools presents a severe indexing challenge.
A standard top-$K$ router computes routing scores for all experts via a projection matrix
$W_g \in \mathbb{R}^{d \times N}$ and set $\mathcal{G}(x)=xW_g$ in Eq.~\ref{eq:gating-score}.
When $N$ reaches millions, the computational cost of $\mathcal{G}(x)$ ($O(Nd)$) and the memory required to store $W_g$ becomes prohibitively expensive and often dominates the total inference latency.

\noindent\textbf{Intuition.} The key insight is that the one-dimensional expert index space of size $N$ can be decomposed into the Cartesian product of two lower-dimensional subspaces.
Therefore, to overcome this bottleneck, we introduce the \textbf{Cartesian Product Router}. 
Rather than scoring all $N$ experts with a single $N$-way classifier, we view an expert id $n$ as a \emph{2D coordinate}
$(i,j)$ on a $N_r \times N_c$ grid (with $N=N_rN_c$).
The router predicts two low-dimensional distributions over \emph{rows} and \emph{columns}, and composes them to score any
expert on the grid. This is analogous in spirit to product-structured indexing (e.g., PKM~\cite{lample2019pkm}):
We replace one prohibitively large projection with two small projections, while still addressing the full $N$-sized expert space.

\noindent\textbf{Implicit Scoring via Factorized Projections.} Our modeling assumption is that the joint probability distribution over the expert grid can be approximated by the product of two independent marginal distributions: 
\begin{equation}
    p(i, j | x) \approx p_r(i | x) \cdot p_c(j | x).
\end{equation}
We replace the single routing matrix $W_g$ with two smaller matrices, $W_r \in \mathbb{R}^{d \times N_r}$ and $W_c \in \mathbb{R}^{d \times N_c}$. For an input token $x$, the row and column logits are computed as 
\begin{equation}
    s_r = x W_r \;, \; s_c = x W_c . \
\end{equation}
Then the log-probabilities for each subspace are obtained via the LogSoftmax function for numerical stability:
\begin{equation}
    p_r = \operatorname{LogSoftmax}(s_r) \;,\; p_c = \operatorname{LogSoftmax}(s_c) .
\end{equation}
Since $p_r$ and $p_c$ are \emph{log}-probabilities (via $\operatorname{LogSoftmax}$), the factorized product becomes additive in log-space: ($\log p(i,j\mid x)\approx p_r[i]+p_c[j]$).
The score for an expert at coordinate $(i, j)$ is the sum of the corresponding log-probabilities, which implicitly defines a score matrix $\mathcal{S} \in \mathbb{R}^{N_r \times N_c}$ without its materialization: 
\begin{equation}
    \label{eq:implicit_mat}
    \mathcal{S}_{ij} = p_r[i] + p_c[j]
\end{equation}

\noindent\textbf{Parallel Top-$K$ Selection.}
Although $\mathcal{S}\in\mathbb{R}^{N_r\times N_c}$ is never materialized, its entries can be computed on-the-fly as described in Eq.~\ref{eq:implicit_mat} from the global vectors $p_r$ and $p_c$.
We therefore partition the implicit grid into tiles and assign each tile to parallel GPU thread blocks.
Each block computes scores for its tile and extracts local top-$K$ candidates. These candidates are then merged via a lightweight reduction to obtain the global top-$K$ expert indices $\mathcal{I}_x$. The routing weights $\mathbf{g}_x$ are computed by normalizing the corresponding top-$K$ scores according to Eq.~\ref{eq:gating-score}.

\noindent\textbf{Complexity Analysis.}
The factorized router reduces the projection cost (and router parameter size) from $O(Nd)$ to $O(\sqrt{N}d)$. Top-$K$ selection is performed on the implicit grid via a tiled GPU search and reduction; although the total score-evaluation work still scales with $N$, it is highly parallel and incurs negligible wall-clock overhead in practice. See Appendix~B for the full derivation and details.

\begin{figure*}[!ht]
    \centering
    \includegraphics[width=\linewidth]{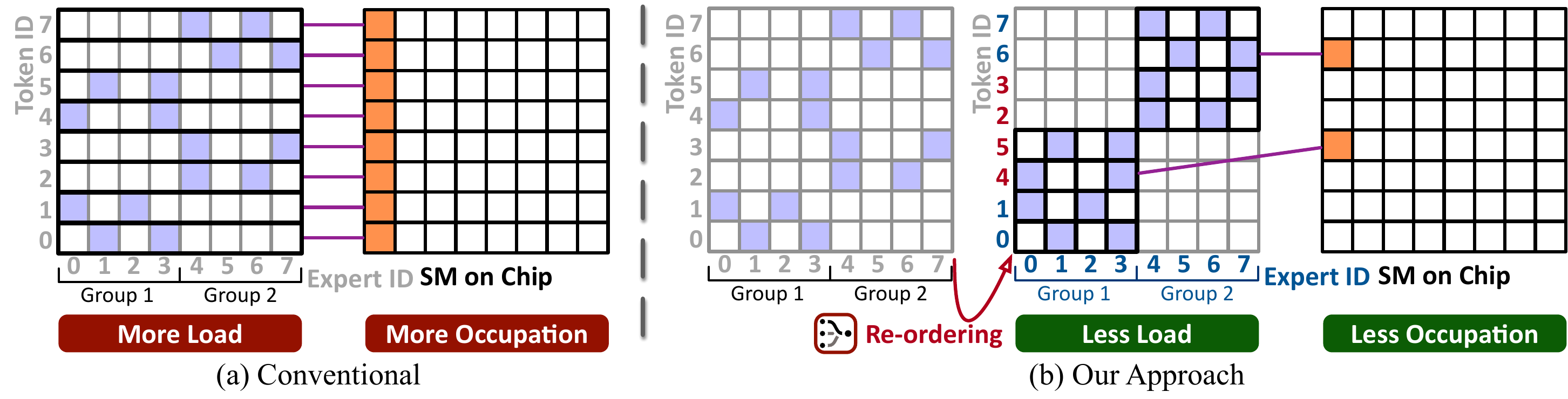}
    \vspace{-.5em}
    \caption{
        \textbf{Comparison of Execution Paradigms: Token-Centric vs. Expert-Centric Scheduling.}
        \textbf{(a) Conventional:} Tokens independently fetch parameters from scattered experts, leading to random memory accesses (high load overhead) and fragmented vector-vector computations that underutilize on-chip SMs.
        \textbf{(b) Our Approach:} We invert the execution order using expert-centric scheduling.
        \textbf{Left-to-Right:} First, tasks are reordered: we compress active experts into dense groups (e.g., experts 0--3 are grouped into Group 1) and sort tasks by Token ID within each group.
        \textbf{Matrix Fusion:} This reorganization allows us to merge individual token-expert pairs into dense tensors.
        Instead of scattered ops, the GPU executes efficient \textbf{Grouped GEMM} kernels (rightmost block), where a block of expert weights is loaded once and reused across stacked tokens, maximizing Tensor Core utilization and memory bandwidth.
        }
    \vspace{-1.0em}
    \label{fig:expert_centric_scheduling}
\end{figure*}

\subsection{Expert-Centric Scheduling}
\label{subsec:expert_centric_scheduling}

The Cartesian Product Router (Section~\ref{subsec:cartesian_product_router}) operates on a per-token basis: for each input token $x$, it efficiently selects the top-$K$ expert indices $\mathcal{I}_{x}$ and computes gating weights $\mathbf{g_x}$. In practice, however, execution processes a \emph{batch} of tokens $\mathcal{X} = \{x_l\}_{l=0}^{L-1}$. While routing is efficient, token-centric execution becomes a bottleneck at batch scale: each token independently fetches its selected expert parameters from HBM via scattered accesses, fragmenting memory traffic and preventing vectorization, thus limiting throughput.

To address this, we propose \textbf{Expert-Centric Scheduling}. Unlike standard static approaches that iterate over all experts, our method dynamically organizes computation based on \emph{active} experts. The pipeline proceeds as follows: we first \emph{collect} routed computation tasks, \emph{group} physically nearby experts, then \emph{reorder} tasks according to these groups to improve locality, and finally \emph{execute} the resulting workloads using high-throughput GEMM kernels.

\noindent\textbf{Task Collection and Active Expert Compression.}
For a batch of $L$ tokens $\mathcal{X}=\{x_l\}_{l=0}^{L-1}$ with top-$K$ routing, the router returns, for each token $l$, an ordered expert index list $\mathcal{I}_l = (I_{l,0}, \ldots, I_{l,K-1})$ and the corresponding gating weights $\mathbf{g}_l = (g_{l,0}, \ldots, g_{l,K-1})$, where $g_{l,k}$ denotes the routing weight assigned to expert $I_{l,k}$. We first flatten these decisions into a list of $M = L \times K$ tasks:
\begin{equation}
\label{eq:task_flatten}
    \tilde{\mathcal{T}} = \big\{(x_l,\, I_{l,k},\, g_{l,k}) \;\big|\; l \in [0, L),\, k \in [0, K)\big\}
\end{equation}
and collect the set of \textbf{unique active experts} $\mathbb{E}_{\text{active}}=\bigcup_{x\in\mathcal{X}}\mathcal{I}_x$, sorted by global expert ID. We then partition this ordered list into \textbf{contiguous groups} of size $B$, so that experts with nearby IDs are placed in the same group.
Specifically, the $\tau$-th expert in $\mathbb{E}_{active}$ is assigned to group $q_\tau = \lfloor \tau/ B \rfloor$.
Consequently, the number of execution groups is determined solely by the active sparsity, i.e., $N_{groups} = \lceil |\mathbb{E}_{active}| / B \rceil$, ensuring that every compute group (except the last) is fully populated.

\noindent\textbf{Hierarchical Sorting.}
We reorganize the tasks $\tilde{\mathcal{T}}$ by performing a hierarchical sort. The primary key is the Group ID $q$, and the secondary key is the Token ID $l$.
\begin{equation}
\mathcal{T} = \operatorname{Sort}(\tilde{\mathcal{T}}, \text{keys}=(q, l))
\end{equation}
This sorting strategy improves hardware efficiency at two levels: (i) \emph{Inter-Group Locality:} tasks targeting the same set of $B$ active experts are clustered together; (ii) \emph{Intra-Group Coalescing:} within each group, processing tasks in increasing order of token ID $l$ improves coalescing for input reads and output scatters.

\noindent\textbf{Grouped GEMM Execution.} After hierarchically sorting the tasks, we process each of the $N_{groups}$ active groups. For a given group $q$, we first gather the corresponding expert parameters into dense blocks $W_q, V_q \in \mathbb{R}^{B \times d}$. Concurrently, we stack the input tokens and gating weights for all tasks assigned to this group, forming a dense input tensor $\mathbf{X}_q \in \mathbb{R}^{T_q \times d}$ and a gating vector $\mathbf{G}_q \in \mathbb{R}^{{T_q} \times B}$, where $T_q$ denotes the number of tasks in group $q$. The entire computation is then performed by a single fused operation:
\begin{equation}
    \mathbf{O}_q = (\mathbf{G}_q \odot \sigma(\mathbf{X}_q W_q^\top)) V_q
\end{equation}
where $\mathbf{O}_q \in \mathbb{R}^{T_q \times d}$ is the output block. The per-task outputs are subsequently written back via scatter-add, preserving the semantics of Eq.~\ref{eq:assembly}.

\noindent\textbf{Why Expert-Centric Scheduling is Efficient.}
As illustrated in Figure~\ref{fig:expert_centric_scheduling}, unlike the token-centric paradigm, our expert-centric approach clusters spatially proximate experts into contiguous groups, reorders tasks by token ID within each group, and fuses the resulting workloads into a small number of high-throughput Grouped GEMM kernels.
By executing tasks in expert-centric order and sorting by token ID within each group, we increase parameter reuse and improve memory locality, which raises effective bandwidth and enables high-throughput Grouped GEMM execution.
A detailed complexity analysis is deferred to Appendix~\ref{app:complexity_analysis}.

\section{Experiments}
\label{sec:experiments}

\subsection{Experimental Setup}

\noindent\textbf{Model Architectures.}
We compare six FFN variants: (i) \textbf{Dense} (standard MLP), (ii) \textbf{Gshard}~\cite{lepikhin2020gshard}, (iii) \textbf{DeepSeekMoE}~\cite{dai2024deepseekmoe}, (iv) \textbf{PKM}~\cite{lample2019pkm}, (v) \textbf{PEER}~\cite{he2024peer}, and (vi) \textbf{\method} (ours). All models adopt Grouped Query Attention (GQA)~\cite{ainslie2023gqa}. For fair comparison, we keep the Transformer backbone identical across methods (depth, width, and attention configuration) and vary only the FFN module. It is important to emphasize that we prioritize \textbf{architectural comparison} via controlled \textbf{pre-training from scratch} rather than comparing against off-the-shelf checkpoints. We define the activated-parameter budget as the number of unique parameters utilized in the forward pass of a single token, including embeddings, attention weights, the shared dense FFN, router projections, and the top-$K$ active MoE experts. For efficiency baselines, we ensure state-of-the-art implementations (see Appendix~\ref{app:explicit_setup} for details).

We evaluate \method under three complementary settings. For \textbf{Speed and Memory Benchmarking}, we fix a \textbf{200M} backbone and sweep the activated-parameter budget (Act Params) and the number of activated tokens (Act Tokens) as summarized in Table~\ref{table:speed_memory_configs} (see Appendix~\ref{app:explicit_setup}). For \textbf{scaling-law} experiments, we train MoE families with \textbf{280M-A80M, 800M-A200M, 2.7B-A680M, and 6.4B-A1.7B} configurations (where A denotes the activated parameter budget) alongside their \textbf{Dense} counterparts with matched activated parameters, using matched backbones and training recipes across baselines (Table~\ref{table:lm_configs} in Appendix~\ref{app:explicit_setup}). For \textbf{downstream evaluation}, we report zero-shot results using the \textbf{6.4B-A1.7B} models.

\noindent\textbf{Training Data and Tokenization.}
Models are pre-trained on the SmolLMCorpus~\cite{allal2025smollm2smolgoesbig}, a high-quality corpus of 40 billion tokens spanning Web, Textbook, Code, and Math domains. This diverse mixture establishes fundamental linguistic proficiency and broad general knowledge. We employ the NeoX tokenizer~\cite{black2022gpt} with a vocabulary size of 128,256 tokens.

\noindent\textbf{Training Strategy and Hyper-Parameters.}
We use the AdamW optimizer~\cite{Loshchilov2017FixingWD} with the WSD learning rate scheduler~\cite{hägele2024scalinglawscomputeoptimaltraining}. Hyperparameters follow optimal scaling laws~\cite{li2025predictablescalei} and Chinchilla compute-optimality protocols~\cite{hoffmann2022empirical}. We run experiments in the NVIDIA PyTorch container~\cite{nv2022pytorch} with Hugging Face Transformers~\cite{wolf-etal-2020-transformers}. All inference/evaluation runs use a single node with 8$\times$ NVIDIA A100 GPUs.

\noindent\textbf{Evaluation Benchmarks.}
We evaluate downstream and reasoning performance with Hugging Face LightEval~\cite{Clémentine2023lighteval} on seven commonsense benchmarks: MMLU~\cite{hendrycks2021measuring} (multitask knowledge), TriviaQA~\cite{joshi2017triviaqa} (factual recall), ARC~\cite{clark2018think} (science reasoning), PIQA~\cite{bisk2020piqa} (physical commonsense), HellaSwag~\cite{zellers2019hellaswag} (commonsense inference), OBQA~\cite{mihaylov2018can} (open-book QA), and Winogrande~\cite{sakaguchi2019winogrande} (coreference resolution); and five extended benchmarks: GSM8K~\cite{cobbe2021trainingverifierssolvemath} (math reasoning), MATH~\cite{hendrycks2021measuringmathematicalproblemsolving} (competition math), BBH~\cite{suzgun2022challengingbigbenchtaskschainofthought} (hard reasoning), MBPP~\cite{austin2021programsynthesislargelanguage} (code generation), and HumanEval~\cite{chen2021codex} (code synthesis).

\subsection{Main Results}
\label{subsec:main_results}

\begin{table*}[!t]
  \centering
  \caption{
    \textbf{Performance on Downstream Benchmarks for 6.4B-A1.7B MoE Models and the 1.7B Dense Baseline.}.
    The best results for each size are in bold, and the second-best results are underlined. For the pre-trained base model, \method performs well on most tasks, demonstrating its effectiveness.
  }
  \resizebox{\linewidth}{!}
  {
  \small

    \begin{tabular}{@{}lcccccccccccc@{}}
    \toprule
    \sc{Model} & \sc{MMLU} & \sc{TriviaQA} & \sc{ARC} & \sc{PIQA} & \sc{HellaSwag} & \sc{OBQA} & \sc{WinoGrande} &\sc{Avg.} \\
    & \sc{acc $\uparrow$} & \sc{acc $\uparrow$} & \sc{acc $\uparrow$} & \sc{acc $\uparrow$} & \sc{acc $\uparrow$} & \sc{acc $\uparrow$} & \sc{acc $\uparrow$} & \sc{avg $\uparrow$} \\
    \midrule
    Dense & 35.4 & 9.4 & 53.4 & 72.9 & 56.1 & 37.0 & 57.3 & 45.9 \\
    Gshard & 36.7 & 16.7 & 58.3 & 75.3 & 59.3 & 38.7 & \underline{59.5} & 49.2 \\
    DeepSeekMoE & 37.1 & \underline{17.4} & \underline{60.7} & \underline{77.2} & \textbf{61.2} & 38.9 & 59.1 & \underline{50.2} \\
    PKM & 36.3 & 12.2 & 53.6 & 73.8 & 52.7 & 38.2 & 56.7 & 46.2 \\
    PEER & \underline{37.4} & 16.9 & 57.4 & 75.9 & 56.3 & \underline{39.1} & 59.4 & 48.9 \\
    \method (ours) & \textbf{37.5} & \textbf{18.5} & \textbf{61.0} & \textbf{78.7} & \underline{60.9} & \textbf{40.3} & \textbf{59.7} & \textbf{50.9} \\
    \bottomrule
    \end{tabular}
  }
  \label{table:downstream_zeroshot}
\end{table*}

\noindent\textbf{Main Results.}
We summarize the main empirical findings of \method from two complementary perspectives: \emph{model quality} under a fixed active-parameter budget, and \emph{system efficiency} (latency/memory) when executing the corresponding routed feed-forward computation.

\noindent\textbf{Downstream Performance.} As shown in Table~\ref{table:downstream_zeroshot}, our 6.4B-A1.7B model achieves the best average zero-shot accuracy (50.9), outperforming both coarse-grained (e.g., +0.7 vs. DeepSeekMoE) and fine-grained (+2.0 vs. PEER) baselines. The results highlight the benefits of our heterogeneous design: compared to coarse-grained models, \method's precision on knowledge-intensive tasks like TriviaQA (+1.1) and OBQA (+1.4) is superior. Conversely, compared to fine-grained models with limited expressivity, \method's shared dense expert boosts performance on reasoning-heavy benchmarks such as ARC (+3.6) and HellaSwag (+4.6).

\begin{table*}[!t]
  \centering
  \caption{
    \textbf{Performance and Throughput on Reasoning Benchmarks for 6.4B-A1.7B MoE Models and the 1.7B Dense Baseline.}
    Throughput is measured with 40 batched parallel requests ($\sim$512 input tokens, $\sim$8192 output tokens) on 8$\times$ A100 GPUs.
    The best results are in bold, and the second-best are underlined.
    For the fine-tuned models, \method performs well on most tasks and achieves the highest throughput.
  }
  \resizebox{\linewidth}{!}
  {
    \begin{tabular}{@{}lccccccccc@{}}
    \toprule
    \sc{Model} & \sc{Throughput} & \sc{MMLU} & \sc{BBH} & \sc{GSM8K} & \sc{MATH} & \sc{ARC-C} & \sc{MBPP} & \sc{HumanEval} & \sc{Avg.} \\
    & (tok/s) & \sc{acc $\uparrow$} & \sc{acc $\uparrow$} & \sc{acc $\uparrow$} & \sc{acc $\uparrow$} & \sc{acc $\uparrow$} & \sc{acc $\uparrow$} & \sc{acc $\uparrow$} & $\uparrow$ \\
    \midrule
    Dense & $\sim$14.8k & 46.4 & 37.7 & 46.3 & 11.7 & 37.4 & 40.0 & 6.7 & 32.3 \\
    Gshard & $\sim$14.0k & 48.2 & 41.5 & 61.9 & 18.9 & 44.1 & 47.4 & 8.5 & 38.6 \\
    DeepSeekMoE & $\sim$13.4k & \underline{49.1} & \underline{43.0} & \underline{65.6} & \underline{21.3} & \underline{45.6} & \underline{49.9} & \underline{9.1} & \underline{40.5} \\
    PKM & $\sim$7.9k & 36.3 & 24.8 & 18.9 & 4.1 & 30.2 & 17.6 & 1.8 & 19.1 \\
    PEER & $\sim$11.6k & 43.8 & 35.6 & 31.4 & 7.9 & 35.9 & 29.7 & 4.9 & 27.0 \\
    \method (ours) & $\sim$14.2k & \textbf{50.2} & \textbf{44.6} & \textbf{69.8} & \textbf{24.1} & \textbf{46.8} & \textbf{52.3} & \textbf{9.8} & \textbf{42.5} \\
    \bottomrule
    \end{tabular}
  }
  \label{table:throughput_reasoning}
\end{table*}

\noindent\textbf{Reasoning Performance and Throughput.} To further validate generalization beyond commonsense tasks, we report results on reasoning benchmarks alongside end-to-end throughput. As shown in Table~\ref{table:throughput_reasoning}, \method achieves the highest throughput among all MoE variants ($\sim$14.2k tok/s), comparable to the Dense baseline ($\sim$14.8k tok/s) and significantly outperforming fine-grained baselines (PEER: $\sim$11.6k tok/s, PKM: $\sim$7.9k tok/s). This confirms that Expert-Centric Scheduling effectively eliminates the memory bandwidth bottleneck at the full-model level.
On the extended benchmarks, \method achieves an average score of 42.5, outperforming DeepSeekMoE (40.5) by +2.0 and PEER (27.0) by +15.5. The large gap over PEER indicates that purely fine-grained designs without a shared dense backbone struggle with sustained multi-step logic, while \method's hybrid architecture effectively combines precise knowledge retrieval with stable reasoning capacity.

\begin{figure}[t!]
    \centering
    \includegraphics[width=0.9\linewidth]{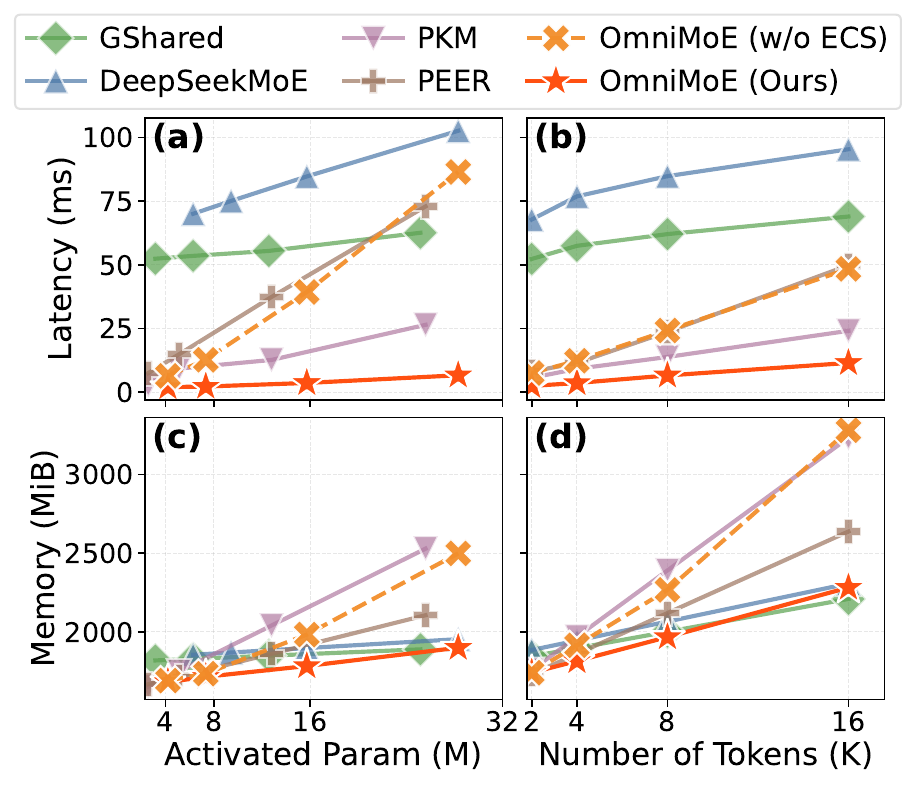}
    \vspace{-.5em}
    \caption{
    \textbf{End-to-End Efficiency Comparison.} (a, b) Inference latency and (c, d) peak memory versus activated parameters (left column) and input token count (right column). Baselines include Dense, Gshard, DeepSeekMoE, PKM, and PEER. OmniMoE achieves consistently lower latency than DeepSeekMoE and fine-grained baselines (PKM/PEER), while maintaining a peak memory footprint comparable to coarse-grained MoEs.
    }
    \vspace{-1.0em}
    \label{fig:comparison}
\end{figure}

\begin{figure}[t!]
    \centering
    \includegraphics[width=0.92\linewidth]{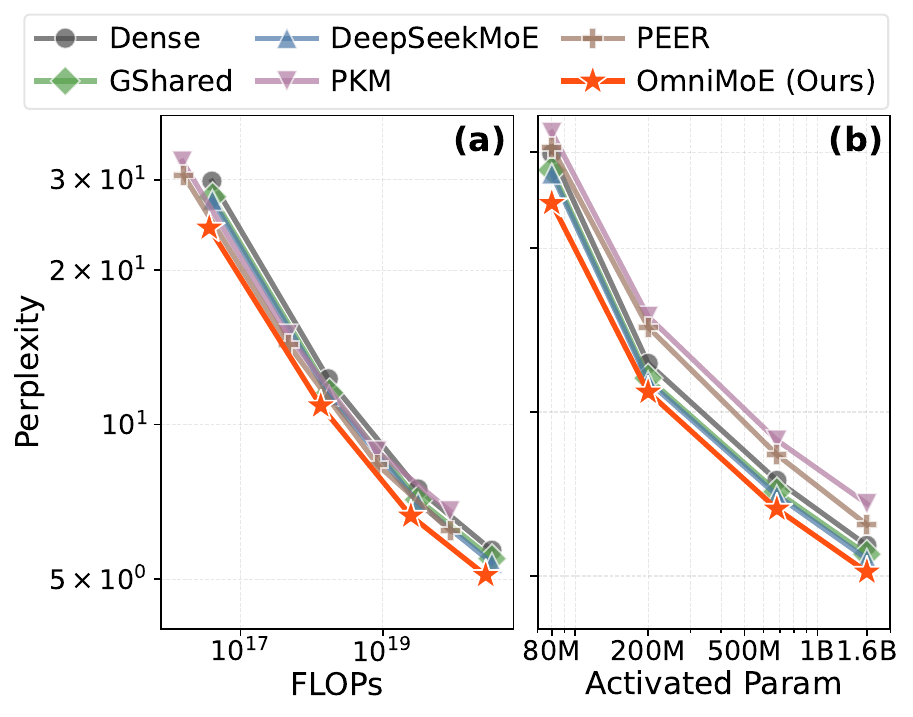}
    \vspace{-.5em}
    \caption{
        \textbf{Scaling Laws.}
        Validation perplexity (lower is better) versus (a) training FLOPs and (b) activated parameters.
        \method consistently outperforms all baselines, achieving the best trade-off between model quality and computational cost.
        }
    \vspace{-1.0em}
    \label{fig:scaling_laws}
\end{figure}

\noindent\textbf{End-to-End Efficiency and Scalability.} Figure~\ref{fig:comparison} demonstrates that \method is substantially more efficient. Despite activating a comparable or even larger number of parameters (\method: 28M, PEER: 26M, DeepSeekMoE: 28M), \method achieves substantially lower latency, reducing inference time from 73 ms (PEER) and 102 ms (DeepSeekMoE) to 6.7 ms at 4,096 tokens, \textbf{10.9$\times$ and 15.2$\times$ speedup} respectively, while maintaining a memory footprint comparable to coarse-grained MoEs. This gain stems directly from Expert-Centric Scheduling, which transforms scattered memory accesses into coalesced, reusable reads, thus shifting the execution from memory-bound to compute-bound.

Interestingly, although DeepSeekMoE uses coarse-grained FFN experts, it can be slower than fine-grained PEER at large token counts or activated budgets. This is largely due to packing/alignment overhead in tiled coarse-grained kernels, where routed tokens must be reordered and padded to fixed block sizes, causing redundant computation and extra memory traffic. In contrast, \method reshapes fine-grained activations into compact, expert-centric batched matrix operations. We report strict end-to-end latency for all methods, including all scheduling/reordering overheads for \method and the corresponding layout-transformation/alignment costs for baselines, confirming net gains from improved hardware utilization. Furthermore, we verify the scalability of \method in distributed training settings (Appendix~\ref{app:ep_communication}). We observe that the communication overhead saturates once the expert pool size exceeds the active token count, demonstrating that \method can scale to millions of experts with constant communication cost.

\noindent\textbf{Scaling laws (perplexity vs. compute/activated parameters).}
Figure~\ref{fig:scaling_laws} compares the scaling behavior of different FFN variants. Under matched training FLOPs, \method consistently achieves the lowest perplexity among all baselines, indicating superior compute efficiency. Moreover, when controlling for the activated parameter budget, \method also attains the lowest perplexity, demonstrating higher parameter efficiency. As the activated capacity increases, \method benefits more steadily from additional experts, reflecting the complementary roles of fine-grained activation for long-tail knowledge and the shared dense MLP branch for stable general reasoning.

\subsection{Ablation Studies}
\label{subsec:ablation_studies}

\begin{table*}[!t]
    \centering
    \caption{
        \textbf{Ablation Study}.
        All metrics are reported relative to the full model. Lower is better for Latency, Memory, PPL, and Unevenness; higher is better for Knowledge Performance, Reasoning Performance, and Expert Usage.
    }
    \resizebox{\linewidth}{!}
    {
        \begin{tabular}{@{}p{0.25\linewidth}ccccccc@{}}
        \toprule
        \sc{Methods} & \sc{Latency} & \sc{Memory} & \sc{PPL} & \sc{Knowledge Perf.} & \sc{Reasoning Perf.} & \sc{Expert Usage} & \sc{Unevenness} \\
        & $\downarrow$ & $\downarrow$ & $\downarrow$ & $\uparrow$ & $\uparrow$ & $\uparrow$ & $\downarrow$ \\
        \midrule
        Full & 1.0\texttimes & 1.0\texttimes & 1.0\texttimes & 1.0\texttimes & 1.0\texttimes & 100\% & 0.24 \\
        w/o Shared Dense MLP & 0.86\texttimes & 0.98\texttimes & 1.2\texttimes & 0.91\texttimes & 0.79\texttimes & 100\% & 0.27 \\
        w/o Cartesian Product Router & 30.6\texttimes & 337.5\texttimes & 1.4\texttimes & 0.66\texttimes & 0.79\texttimes & 4\% & 0.77 \\
        w/o Expert-Centric Scheduling & 24.8\texttimes & 417.7\texttimes & {1.0\texttimes} & 1.0\texttimes & 1.0\texttimes & 100\% & 0.24 \\
        \bottomrule
        \end{tabular}
    }
    \vspace{-0.5em}
    \label{table:ablation}
\end{table*}

We ablate the three core components of \method to isolate their individual contributions.
Table~\ref{table:ablation} isolates the impact of three core components in \method: (i) the \textbf{Shared Dense MLP} for general stability, (ii) the \textbf{Cartesian Product Router} for routing quality, and (iii) \textbf{Expert-Centric Scheduling} for system efficiency. All metrics are normalized to the full model (lower is better for Latency/Memory/PPL/Unevenness).
To characterize expert utilization, we follow PKM~\cite{lample2019pkm} and PEER~\cite{he2024peer} and report two distribution metrics based on the normalized expert retrieval frequency $z \in \mathbb{R}^{N}$:
\vspace{-.5em}
\begin{itemize}
    \setlength\itemsep{0em}
    \item \textbf{Expert Usage}: The fraction of experts activated at least once, defined as $\frac{1}{N} |\{i \mid z_i > 0\}|$.
    \item \textbf{Unevenness}: The KL divergence from a uniform distribution, computed as $D_{\text{KL}}(z \| \mathcal{U}) = \sum_i z_i \log(N z_i)$, where lower values indicate more balanced load.
\end{itemize}

\noindent\textbf{Effect of the shared dense MLP.} Removing the shared dense MLP slightly improves efficiency (0.86$\times$ latency, 0.98$\times$ memory) but hurts both perplexity (1.2$\times$) and downstream performance (0.91$\times$ knowledge and 0.79$\times$ reasoning).This suggests that the shared dense branch serves as a critical foundational backbone complementary to fine-grained retrieval. It handles common linguistic patterns and reasoning steps, allowing the routed branch to focus exclusively on fetching token-specific long-tail knowledge.

\noindent\textbf{Effect of the Cartesian Product Router.} Replacing the Cartesian Product Router with a standard dense routing projection 
leads to a dramatic efficiency regression (30.6$\times$ latency and 337.5$\times$ memory), driven by the cost of computing and storing full-dimension logits. Crucially, it also degrades model quality (1.4$\times$ PPL). Notably, \emph{expert usage} collapses to only 4\% and unevenness increases from 0.24 to 0.77.
Despite rigorous tuning of the standard auxiliary load-balancing loss for this baseline during training, the naive gate fails to learn distinct specializations over the massive expert space, collapsing into a few dominant experts.

\noindent\textbf{Effect of Expert-Centric Scheduling.} Reverting Expert-Centric Scheduling to the standard token-centric execution mechanism preserves quality (metrics remain at 1.0$\times$) but incurs a massive system cost (24.8$\times$ latency and 417.7$\times$ memory). The peak memory increase corresponds to the materialization of full routing tensors required by the standard baseline, which our scheduling avoids. This confirms that our scheduling strategy is the primary source of acceleration: by inverting the loop order to process experts sequentially, we transform strictly random HBM accesses into streaming reads and maximize on-chip tensor reuse, eliminating the memory bandwidth bottleneck inherent to fine-grained MoEs.

\begin{figure}[!t]
    \centering
    \includegraphics[width=\linewidth]{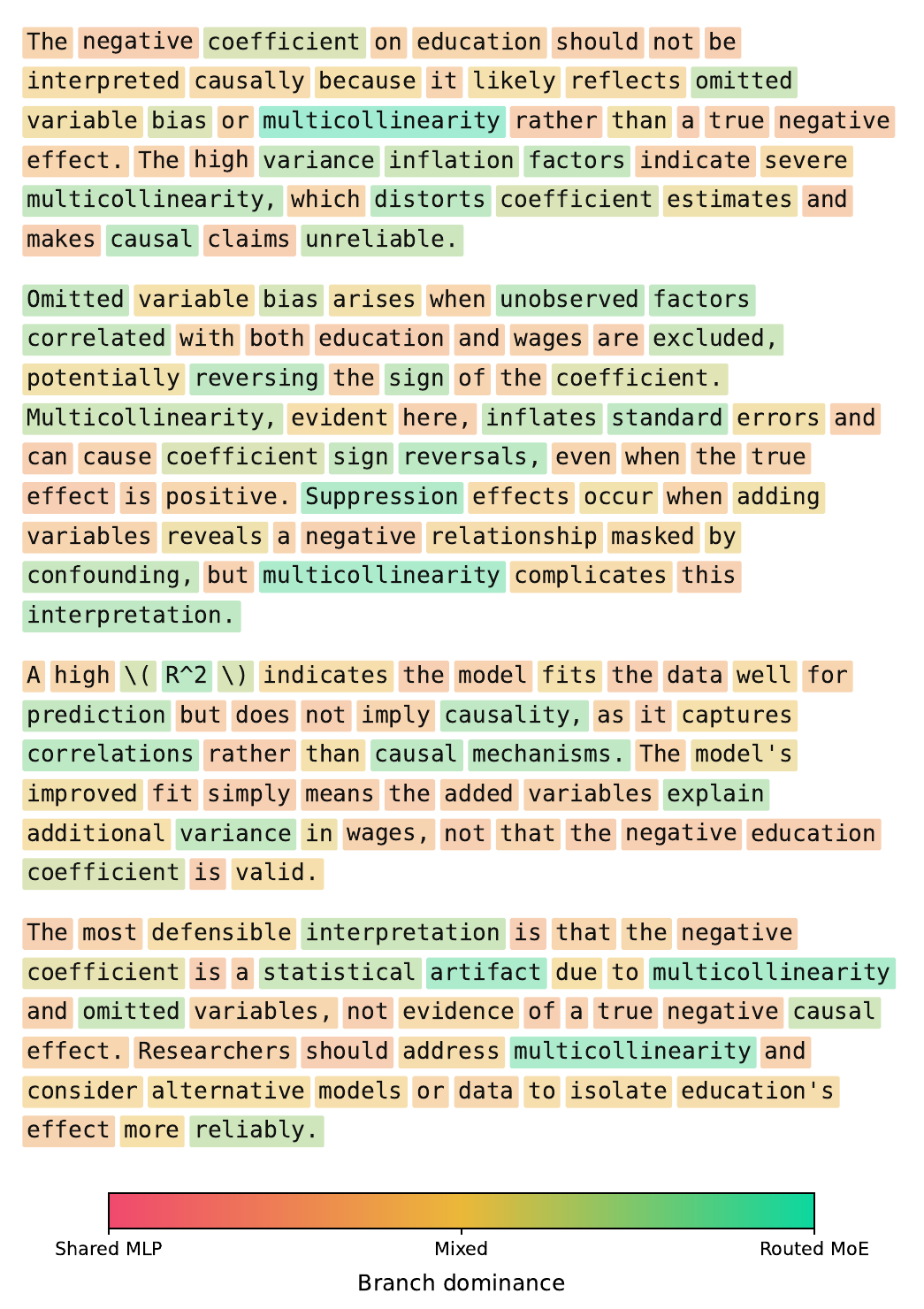}
    \vspace{-.5em}
    \caption{
        \textbf{Token-Level Branch Dominance Analysis.}
        Each token is colored by which branch contributes more to the final prediction: the Shared MLP (red) or the Routed MoE (green). Factual entities and rare tokens are predominantly served by the routed atomic experts, while common linguistic patterns and reasoning connectives are handled by the shared dense MLP, empirically validating the functional specialization of the two branches.
    }
    \vspace{-2.0em}
    \label{fig:token_case_study}
\end{figure}

\noindent\textbf{Branch Specialization Analysis.} To directly validate the division of labor between the two branches, we visualize per-token branch dominance in Figure~\ref{fig:token_case_study}. For each token, we measure the relative contribution of the shared dense MLP versus the routed atomic experts to the final hidden state. The results reveal a clear functional specialization: factual entities and domain-specific terms are predominantly served by the routed branch, while common function words and reasoning connectives rely on the shared MLP. This pattern is consistent with the ablation findings in Table~\ref{table:ablation}, where removing the shared MLP disproportionately hurts reasoning performance while preserving expert usage, confirming that the two branches fulfill complementary roles.

\section{Related Work}
\label{sec:related_work}

\noindent\textbf{MoE Architectures.}
Conditional computation via Mixture-of-Experts (MoE) enables scaling model capacity with bounded per-token cost~\cite{shazeer17outrageous, sun2025speedsurvey, mu2025surveymoe,liu2025surveyinferenceopt}. Early MoE Transformers predominantly adopt coarse-grained FFN experts with lightweight routing, exemplified by Switch Transformers’ Top-1 gating~\cite{fedus22switch} and GShard~\cite{lepikhin2020gshard}. Motivated by scaling-law evidence for improved specialization, recent work trends toward finer-grained expert designs~\cite{krajewski2024scalinglaw, tian2025moescalinglaw}; modern LLMs (e.g., DeepSeek-V3~\cite{deepseek2025deepseekv3}, KIMI-K2~\cite{kimi2025k2}) scale to hundreds of experts and often include shared experts to stabilize general knowledge~\cite{dai2024deepseekmoe}. Pushing granularity to the extreme, PKM~\cite{lample2019pkm} and PEER~\cite{he2024peer} replace FFNs with million-scale tiny experts (e.g., embeddings), improving routing precision but reducing per-expert expressivity. Additionally, prior work reports redundant activation in FFNs/MoEs~\cite{li2023lazyneuron, szatkowski2024d2dmoe, zhou2025dern, yang2024xmoe}; MoNE~\cite{cheng2025mone} prunes computation within activated coarse-grained experts but retains expert-level top-$K$ routing, whereas \method routes over atomic experts to enable finer-grained control of activated parameters.

\noindent\textbf{Efficient MoE Systems.} System optimizations for MoE generally focus on kernel fusion and communication scheduling for coarse-grained experts. Frameworks such as DeepSpeed-MoE~\cite{rajbhandari2022deepspeed}, Fast-MoE~\cite{he2021fastmoe}, and MegaBlocks~\cite{gale2023meagblocks} optimize GEMM kernels and handle variable-length sequences to mitigate padding overheads in coarse-grained MoEs, whereas our Expert-Centric Scheduling targets fine-grained atomic experts, converting scattered memory accesses into contiguous batched operations.
Recent work like SonicMoE~\cite{guo2025sonicmoe} improves efficiency with memory-efficient algorithms, minimal activation caching, and tile-aware token rounding to reduce padding waste in Grouped GEMM kernels.
Other works, including PIT~\cite{zheng2023pit} and ScatterMoE~\cite{tan2024scattermoe}, further exploit dynamic sparsity to prune invalid computations within activated coarse-grained experts but retain expert-level top-$K$ routing.
In contrast, \method introduces Expert-Centric Scheduling to transform scattered memory accesses into hardware-efficient batched operations.
Concurrent works Klotski~\cite{fang2025klotskiefficientmixtureofexpertinference} and ExpertFlow~\cite{he2026expertflowefficientmixtureofexpertsinference} also batch tokens per expert but target inter-request scheduling and cross-device load balancing for deployed coarse-grained MoEs; our scheduling instead operates intra-batch on a single device, addressing the scattered HBM reads unique to million-scale atomic experts.

\section{Conclusion}
\label{sec:conclusion}

In this paper, we presented \textbf{\method}, a system-algorithm co-designed MoE framework that integrates a shared dense MLP for general-purpose reasoning with massive atomic experts for long-tail knowledge retrieval, thereby enabling more precise parameter activation.
To make large-scale expert activation practical, OmniMoE orchestrates the activation of atomic experts via two key innovations: the \textbf{Cartesian Product Router} and \textbf{Expert-Centric Scheduling}.
Together, these components yield a dramatic $10.9\times$ inference speedup over the state-of-the-art fine-grained baseline, PEER. Moreover, under comparable activated-parameter budgets, \method consistently improves average accuracy and outperforms strong baselines on most benchmarks. These results demonstrate that, with holistic co-design, massive-scale fine-grained MoEs can be both accurate and highly efficient.
Our current implementation relies on custom Triton kernels optimized for NVIDIA GPUs; portability to other hardware backends remains unexplored. Additionally, scaling behavior beyond 6.4B total parameters has not yet been validated empirically and is left for future work.

\section*{Acknowledgements}
This paper was supported by the NSF of China (62402409); Youth S\&T Talent Support Programme of Guangdong Provincial Association for Science and Technology (SKXRC2025461); the Young Talent Support Project of Guangzhou Association for Science and Technology (QT-2025-001); Guangzhou Basic and Applied Basic Research Foundation (2026A1515010269, 2025A04J3935, 2023A1515110545); and Guangzhou-HKUST(GZ) Joint Funding Program (2025A03J3714).
We gratefully acknowledge the FlagOS open-source community and the OpenSeek project for providing the computational resources essential to this work.

\section*{Impact Statement}

This paper presents work whose goal is to advance the field of Machine
Learning. There are many potential societal consequences of our work, none which we feel must be specifically highlighted here.

\bibliography{biblio}

@article{dai2024deepseekmoe,
title={Deepseekmoe: Towards ultimate expert specialization in mixture-of-experts language models},
author={Dai, Damai and Deng, Chengqi and Zhao, Chenggang and Xu, RX and Gao, Huazuo and Chen, Deli and Li, Jiashi and Zeng, Wangding and Yu, Xingkai and Wu, Yu and others},
journal={arXiv preprint arXiv:2401.06066},
year={2024}
}

@article{hägele2024scalinglawscomputeoptimaltraining,
title={Scaling laws and compute-optimal training beyond fixed training durations},
author={H{\"a}gele, Alex and Bakouch, Elie and Kosson, Atli and Von Werra, Leandro and Jaggi, Martin and others},
journal={Advances in Neural Information Processing Systems},
volume={37},
pages={76232--76264},
year={2024}
}

@article{allal2025smollm2smolgoesbig,
title={SmolLM2: When Smol Goes Big--Data-Centric Training of a Small Language Model},
author={Allal, Loubna Ben and Lozhkov, Anton and Bakouch, Elie and Bl{\'a}zquez, Gabriel Mart{\'\i}n and Penedo, Guilherme and Tunstall, Lewis and Marafioti, Andr{\'e}s and Kydl{\'\i}{\v{c}}ek, Hynek and Lajar{\'\i}n, Agust{\'\i}n Piqueres and Srivastav, Vaibhav and others},
journal={arXiv preprint arXiv:2502.02737},
year={2025}
}

@inproceedings{lample2019pkm,
title={Large Memory Layers with Product Keys.},
author={Guillaume Lample and Alexandre Sablayrolles and Marc'Aurelio Ranzato and Ludovic Denoyer and Hervé Jégou},
year={2019},
cdate={1546300800000},
pages={8546-8557},
booktitle={Advances in Neural Information Processing Systems},
}

@inproceedings{hendrycks2021measuring,
title={Measuring Massive Multitask Language Understanding}, 
author={Dan Hendrycks and Collin Burns and Steven Basart and Andy Zou and Mantas Mazeika and Dawn Song and Jacob Steinhardt},
booktitle={International Conference on Learning Representations},
year={2021},
}

@misc{joshi2017triviaqa,
title={TriviaQA: A Large Scale Distantly Supervised Challenge Dataset for Reading Comprehension},
author={Mandar Joshi and Eunsol Choi and Daniel S. Weld and Luke Zettlemoyer},
year={2017},
eprint={1705.03551},
archivePrefix={arXiv},
primaryClass={cs.CL}
}

@article{clark2018think,
title={Think you have Solved Question Answering? Try {A}{R}{C}, the {A}{I}2 Reasoning Challenge},
author={Clark, Peter and Cowhey, Isaac and Etzioni, Oren and Khot, Tushar and Sabharwal, Ashish and Schoenick, Carissa and Tafjord, Oyvind},
journal={arXiv preprint arXiv:1803.05457},
year={2018}
}

@inproceedings{bisk2020piqa,
title={P{I}{Q}{A}: Reasoning about Physical Commonsense in Natural Language},
author={Bisk, Yonatan and Zellers, Rowan and Gao, Jianfeng and Choi, Yejin and others},
booktitle={Proceedings of the AAAI conference on Artificial Intelligence},
volume={34},
year={2020}
}

@inproceedings{zellers2019hellaswag,
title={Hella{S}wag: Can a Machine Really Finish Your Sentence?},
author={Zellers, Rowan and Holtzman, Ari and Bisk, Yonatan and Farhadi, Ali and Choi, Yejin},
booktitle ={Proceedings of the 57th Annual Meeting of the Association for Computational Linguistics},
year={2019}
}

@article{mihaylov2018can,
title={Can a Suit of Armor Conduct Electricity? A New Dataset for Open Book Question Answering},
author={Mihaylov, Todor and Clark, Peter and Khot, Tushar and Sabharwal, Ashish},
journal={arXiv preprint arXiv:1809.02789},
year={2018}
}

@misc{sakaguchi2019winogrande,
title={WinoGrande: An Adversarial Winograd Schema Challenge at Scale}, 
author={Keisuke Sakaguchi and Ronan Le Bras and Chandra Bhagavatula and Yejin Choi},
year={2019},
eprint={1907.10641},
archivePrefix={arXiv},
primaryClass={cs.CL},
url={https://arxiv.org/abs/1907.10641}, 
}

@article{suzgun2022challengingbigbenchtaskschainofthought,
title={Challenging big-bench tasks and whether chain-of-thought can solve them},
author={Suzgun, Mirac and Scales, Nathan and Sch{\"a}rli, Nathanael and Gehrmann, Sebastian and Tay, Yi and Chung, Hyung Won and Chowdhery, Aakanksha and Le, Quoc V and Chi, Ed H and Zhou, Denny and others},
journal={arXiv preprint arXiv:2210.09261},
year={2022}
}

@inproceedings{wolf-etal-2020-transformers,
title="Transformers: State-of-the-Art Natural Language Processing",
author="Thomas Wolf and Lysandre Debut and Victor Sanh and Julien Chaumond and Clement Delangue and Anthony Moi and Pierric Cistac and Tim Rault and Rémi Louf and Morgan Funtowicz and Joe Davison and Sam Shleifer and Patrick von Platen and Clara Ma and Yacine Jernite and Julien Plu and Canwen Xu and Teven Le Scao and Sylvain Gugger and Mariama Drame and Quentin Lhoest and Alexander M. Rush",
booktitle="Proceedings of the 2020 Conference on Empirical Methods in Natural Language Processing: System Demonstrations",
month=oct,
year="2020",
address="Online",
publisher="Association for Computational Linguistics",
pages="38--45"
}

@misc{Clémentine2023lighteval,
author={Fourrier, Clémentine and Habib, Nathan and Kydlíček, Hynek and Wolf, Thomas and Tunstall, Lewis},
title={LightEval: A lightweight framework for LLM evaluation},
year={2023},
version={0.7.0},
url={https://github.com/huggingface/lighteval}
}

@misc{qwen32025,
title={Qwen3},
url={https://qwenlm.github.io/blog/qwen3},
author={Qwen Team},
month={April},
year={2025}
}

@misc{nv2022pytorch,
title={PyTorch Container Image},
author={NVIDIA, Meta},
howpublished={\url{https://catalog.ngc.nvidia.com/orgs/nvidia/containers/pytorch}},
year={2022}
}

@article{black2022gpt,
title={Gpt-NeoX-20B: An Open-source Autoregressive Language Model},
author={Black, Sid and Biderman, Stella and Hallahan, Eric and Anthony, Quentin and Gao, Leo and Golding, Laurence and He, Horace and Leahy, Connor and McDonell, Kyle and Phang, Jason and others},
journal={arXiv preprint arXiv:2204.06745},
year={2022}
}

@article{Loshchilov2017FixingWD,
title={Fixing Weight Decay Regularization in Adam},
author={Ilya Loshchilov and Frank Hutter},
journal={ArXiv},
year={2017},
volume={abs/1711.05101},
url={https://api.semanticscholar.org/CorpusID:3312944}
}

@misc{li2025predictablescalei,
title={Predictable Scale: Part I -- Optimal Hyperparameter Scaling Law in Large Language Model Pretraining}, 
author={Houyi Li and Wenzhen Zheng and Qiufeng Wang and Hanshan Zhang and Zili Wang and Shijie Xuyang and Yuantao Fan and Shuigeng Zhou and Xiangyu Zhang and Daxin Jiang},
year={2025},
eprint={2503.04715},
archivePrefix={arXiv},
primaryClass={cs.LG},
url={https://arxiv.org/abs/2503.04715}, 
}

@article{hoffmann2022empirical,
title={An Empirical Analysis of Compute-Optimal Large Language Model Training},
author={Hoffmann, Jordan and Borgeaud, Sebastian and Mensch, Arthur and Buchatskaya, Elena and Cai, Trevor and Rutherford, Eliza and de Las Casas, Diego and Hendricks, Lisa Anne and Welbl, Johannes and Clark, Aidan and others},
journal={Advances in Neural Information Processing Systems (NeurIPS)},
volume={35},
pages={30016--30030},
year={2022}
}

@article{noam2020swiglu,
  author       = {Noam Shazeer},
  title        = {{GLU} Variants Improve Transformer},
  journal      = {CoRR},
  volume       = {abs/2002.05202},
  year         = {2020},
  url          = {https://arxiv.org/abs/2002.05202},
  eprinttype    = {arXiv},
  eprint       = {2002.05202},
  bibsource    = {dblp computer science bibliography, https://dblp.org}
}

@article{jiang2024mixtral,
  title={Mixtral of experts},
  author={Jiang, Albert Q and Sablayrolles, Alexandre and Roux, Antoine and Mensch, Arthur and Savary, Blanche and Bamford, Chris and Chaplot, Devendra Singh and Casas, Diego de las and Hanna, Emma Bou and Bressand, Florian and others},
  journal={arXiv preprint arXiv:2401.04088},
  year={2024}
}

@article{zoph2022st,
  title={St-moe: Designing stable and transferable sparse expert models},
  author={Zoph, Barret and Bello, Irwan and Kumar, Sameer and Du, Nan and Huang, Yanping and Dean, Jeff and Shazeer, Noam and Fedus, William},
  journal={arXiv preprint arXiv:2202.08906},
  year={2022}
}

@inproceedings{szatkowski2024d2dmoe,
  author       = {Filip Szatkowski and
                  Bartosz W{\'{o}}jcik and
                  Mikolaj Pi{\'{o}}rczynski and
                  Simone Scardapane},
  editor       = {Amir Globersons and
                  Lester Mackey and
                  Danielle Belgrave and
                  Angela Fan and
                  Ulrich Paquet and
                  Jakub M. Tomczak and
                  Cheng Zhang},
  title        = {Exploiting Activation Sparsity with Dense to Dynamic-k Mixture-of-Experts
                  Conversion},
  booktitle    = {Advances in Neural Information Processing Systems 38: Annual Conference
                  on Neural Information Processing Systems 2024, NeurIPS 2024, Vancouver,
                  BC, Canada, December 10 - 15, 2024},
  year         = {2024},
  bibsource    = {dblp computer science bibliography, https://dblp.org}
}

@inproceedings{du2022glam,
  author       = {Nan Du and
                  Yanping Huang and
                  Andrew M. Dai and
                  Simon Tong and
                  Dmitry Lepikhin and
                  Yuanzhong Xu and
                  Maxim Krikun and
                  Yanqi Zhou and
                  Adams Wei Yu and
                  Orhan Firat and
                  Barret Zoph and
                  Liam Fedus and
                  Maarten P. Bosma and
                  Zongwei Zhou and
                  Tao Wang and
                  Yu Emma Wang and
                  Kellie Webster and
                  Marie Pellat and
                  Kevin Robinson and
                  Kathleen S. Meier{-}Hellstern and
                  Toju Duke and
                  Lucas Dixon and
                  Kun Zhang and
                  Quoc V. Le and
                  Yonghui Wu and
                  Zhifeng Chen and
                  Claire Cui},
  title        = {GLaM: Efficient Scaling of Language Models with Mixture-of-Experts},
  booktitle    = {International Conference on Machine Learning, {ICML} 2022, 17-23 July
                  2022, Baltimore, Maryland, {USA}},
  series       = {Proceedings of Machine Learning Research},
  volume       = {162},
  pages        = {5547--5569},
  publisher    = {{PMLR}},
  year         = {2022},
  url          = {https://proceedings.mlr.press/v162/du22c.html},
  bibsource    = {dblp computer science bibliography, https://dblp.org}
}

@inproceedings{krajewski2024scalinglaw,
  author       = {Jan Ludziejewski and
                  Jakub Krajewski and
                  Kamil Adamczewski and
                  Maciej Pi{\'{o}}ro and
                  Michal Krutul and
                  Szymon Antoniak and
                  Kamil Ciebiera and
                  Krystian Kr{\'{o}}l and
                  Tomasz Odrzyg{\'{o}}zdz and
                  Piotr Sankowski and
                  Marek Cygan and
                  Sebastian Jaszczur},
  title        = {Scaling Laws for Fine-Grained Mixture of Experts},
  booktitle    = {Forty-first International Conference on Machine Learning, {ICML} 2024,
                  Vienna, Austria, July 21-27, 2024},
  publisher    = {OpenReview.net},
  year         = {2024},
  url          = {https://openreview.net/forum?id=yoqdlynCRs},
  bibsource    = {dblp computer science bibliography, https://dblp.org}
}

@inproceedings{clark2022scalinglaw,
  author       = {Aidan Clark and
                  Diego de Las Casas and
                  Aurelia Guy and
                  Arthur Mensch and
                  Michela Paganini and
                  Jordan Hoffmann and
                  Bogdan Damoc and
                  Blake A. Hechtman and
                  Trevor Cai and
                  Sebastian Borgeaud and
                  George van den Driessche and
                  Eliza Rutherford and
                  Tom Hennigan and
                  Matthew J. Johnson and
                  Albin Cassirer and
                  Chris Jones and
                  Elena Buchatskaya and
                  David Budden and
                  Laurent Sifre and
                  Simon Osindero and
                  Oriol Vinyals and
                  Marc'Aurelio Ranzato and
                  Jack W. Rae and
                  Erich Elsen and
                  Koray Kavukcuoglu and
                  Karen Simonyan},
  title        = {Unified Scaling Laws for Routed Language Models},
  booktitle    = {International Conference on Machine Learning, {ICML} 2022, 17-23 July
                  2022, Baltimore, Maryland, {USA}},
  series       = {Proceedings of Machine Learning Research},
  volume       = {162},
  pages        = {4057--4086},
  publisher    = {{PMLR}},
  year         = {2022},
  url          = {https://proceedings.mlr.press/v162/clark22a.html},
  bibsource    = {dblp computer science bibliography, https://dblp.org}
}

@misc{tian2025moescalinglaw,
  title = {Towards {{Greater Leverage}}: {{Scaling Laws}} for {{Efficient Mixture-of-Experts Language Models}}},
  shorttitle = {Towards {{Greater Leverage}}},
  author = {Tian, Changxin and Chen, Kunlong and Liu, Jia and Liu, Ziqi and Zhang, Zhiqiang and Zhou, Jun},
  year = 2025,
  month = oct,
  number = {arXiv:2507.17702},
  eprint = {2507.17702},
  primaryclass = {cs},
  publisher = {arXiv},
  doi = {10.48550/arXiv.2507.17702},
  archiveprefix = {arXiv}
}

@inproceedings{lepikhin2020gshard,
  author       = {Dmitry Lepikhin and
                  HyoukJoong Lee and
                  Yuanzhong Xu and
                  Dehao Chen and
                  Orhan Firat and
                  Yanping Huang and
                  Maxim Krikun and
                  Noam Shazeer and
                  Zhifeng Chen},
  title        = {GShard: Scaling Giant Models with Conditional Computation and Automatic
                  Sharding},
  booktitle    = {9th International Conference on Learning Representations, {ICLR} 2021,
                  Virtual Event, Austria, May 3-7, 2021},
  publisher    = {OpenReview.net},
  year         = {2021},
  url          = {https://openreview.net/forum?id=qrwe7XHTmYb},
  bibsource    = {dblp computer science bibliography, https://dblp.org}
}

@misc{kimi2025k2,
  title = {Kimi {{K2}}: {{Open Agentic Intelligence}}},
  shorttitle = {Kimi {{K2}}},
  author = {Team, Kimi and Bai, Yifan and Bao, Yiping and Chen, Guanduo and Chen, Jiahao and Chen, Ningxin and Chen, Ruijue and Chen, Yanru and Chen, Yuankun and Chen, Yutian and Chen, Zhuofu and Cui, Jialei and Ding, Hao and Dong, Mengnan and Du, Angang and Du, Chenzhuang and Du, Dikang and Du, Yulun and Fan, Yu and Feng, Yichen and Fu, Kelin and Gao, Bofei and Gao, Hongcheng and Gao, Peizhong and Gao, Tong and Gu, Xinran and Guan, Longyu and Guo, Haiqing and Guo, Jianhang and Hu, Hao and Hao, Xiaoru and He, Tianhong and He, Weiran and He, Wenyang and Hong, Chao and Hu, Yangyang and Hu, Zhenxing and Huang, Weixiao and Huang, Zhiqi and Huang, Zihao and Jiang, Tao and Jiang, Zhejun and Jin, Xinyi and Kang, Yongsheng and Lai, Guokun and Li, Cheng and Li, Fang and Li, Haoyang and Li, Ming and Li, Wentao and Li, Yanhao and Li, Yiwei and Li, Zhaowei and Li, Zheming and Lin, Hongzhan and Lin, Xiaohan and Lin, Zongyu and Liu, Chengyin and Liu, Chenyu and Liu, Hongzhang and Liu, Jingyuan and Liu, Junqi and Liu, Liang and Liu, Shaowei and Liu, T. Y. and Liu, Tianwei and Liu, Weizhou and Liu, Yangyang and Liu, Yibo and Liu, Yiping and Liu, Yue and Liu, Zhengying and Lu, Enzhe and Lu, Lijun and Ma, Shengling and Ma, Xinyu and Ma, Yingwei and Mao, Shaoguang and Mei, Jie and Men, Xin and Miao, Yibo and Pan, Siyuan and Peng, Yebo and Qin, Ruoyu and Qu, Bowen and Shang, Zeyu and Shi, Lidong and Shi, Shengyuan and Song, Feifan and Su, Jianlin and Su, Zhengyuan and Sun, Xinjie and Sung, Flood and Tang, Heyi and Tao, Jiawen and Teng, Qifeng and Wang, Chensi and Wang, Dinglu and Wang, Feng and Wang, Haiming and Wang, Jianzhou and Wang, Jiaxing and Wang, Jinhong and Wang, Shengjie and Wang, Shuyi and Wang, Yao and Wang, Yejie and Wang, Yiqin and Wang, Yuxin and Wang, Yuzhi and Wang, Zhaoji and Wang, Zhengtao and Wang, Zhexu and Wei, Chu and Wei, Qianqian and Wu, Wenhao and Wu, Xingzhe and Wu, Yuxin and Xiao, Chenjun and Xie, Xiaotong and Xiong, Weimin and Xu, Boyu and Xu, Jing and Xu, Jinjing and Xu, L. H. and Xu, Lin and Xu, Suting and Xu, Weixin and Xu, Xinran and Xu, Yangchuan and Xu, Ziyao and Yan, Junjie and Yan, Yuzi and Yang, Xiaofei and Yang, Ying and Yang, Zhen and Yang, Zhilin and Yang, Zonghan and Yao, Haotian and Yao, Xingcheng and Ye, Wenjie and Ye, Zhuorui and Yin, Bohong and Yu, Longhui and Yuan, Enming and Yuan, Hongbang and Yuan, Mengjie and Zhan, Haobing and Zhang, Dehao and Zhang, Hao and Zhang, Wanlu and Zhang, Xiaobin and Zhang, Yangkun and Zhang, Yizhi and Zhang, Yongting and Zhang, Yu and Zhang, Yutao and Zhang, Yutong and Zhang, Zheng and Zhao, Haotian and Zhao, Yikai and Zheng, Huabin and Zheng, Shaojie and Zhou, Jianren and Zhou, Xinyu and Zhou, Zaida and Zhu, Zhen and Zhuang, Weiyu and Zu, Xinxing},
  year = 2025,
  month = jul,
  number = {arXiv:2507.20534},
  eprint = {2507.20534},
  primaryclass = {cs},
  publisher = {arXiv},
  doi = {10.48550/arXiv.2507.20534},
  urldate = {2026-01-21},
  archiveprefix = {arXiv},
}

@misc{deepseek2025deepseekv3,
  title = {{{DeepSeek-V3 Technical Report}}},
  author = {{DeepSeek-AI} and Liu, Aixin and Feng, Bei and Xue, Bing and Wang, Bingxuan and Wu, Bochao and Lu, Chengda and Zhao, Chenggang and Deng, Chengqi and Zhang, Chenyu and Ruan, Chong and Dai, Damai and Guo, Daya and Yang, Dejian and Chen, Deli and Ji, Dongjie and Li, Erhang and Lin, Fangyun and Dai, Fucong and Luo, Fuli and Hao, Guangbo and Chen, Guanting and Li, Guowei and Zhang, H. and Bao, Han and Xu, Hanwei and Wang, Haocheng and Zhang, Haowei and Ding, Honghui and Xin, Huajian and Gao, Huazuo and Li, Hui and Qu, Hui and Cai, J. L. and Liang, Jian and Guo, Jianzhong and Ni, Jiaqi and Li, Jiashi and Wang, Jiawei and Chen, Jin and Chen, Jingchang and Yuan, Jingyang and Qiu, Junjie and Li, Junlong and Song, Junxiao and Dong, Kai and Hu, Kai and Gao, Kaige and Guan, Kang and Huang, Kexin and Yu, Kuai and Wang, Lean and Zhang, Lecong and Xu, Lei and Xia, Leyi and Zhao, Liang and Wang, Litong and Zhang, Liyue and Li, Meng and Wang, Miaojun and Zhang, Mingchuan and Zhang, Minghua and Tang, Minghui and Li, Mingming and Tian, Ning and Huang, Panpan and Wang, Peiyi and Zhang, Peng and Wang, Qiancheng and Zhu, Qihao and Chen, Qinyu and Du, Qiushi and Chen, R. J. and Jin, R. L. and Ge, Ruiqi and Zhang, Ruisong and Pan, Ruizhe and Wang, Runji and Xu, Runxin and Zhang, Ruoyu and Chen, Ruyi and Li, S. S. and Lu, Shanghao and Zhou, Shangyan and Chen, Shanhuang and Wu, Shaoqing and Ye, Shengfeng and Ye, Shengfeng and Ma, Shirong and Wang, Shiyu and Zhou, Shuang and Yu, Shuiping and Zhou, Shunfeng and Pan, Shuting and Wang, T. and Yun, Tao and Pei, Tian and Sun, Tianyu and Xiao, W. L. and Zeng, Wangding and Zhao, Wanjia and An, Wei and Liu, Wen and Liang, Wenfeng and Gao, Wenjun and Yu, Wenqin and Zhang, Wentao and Li, X. Q. and Jin, Xiangyue and Wang, Xianzu and Bi, Xiao and Liu, Xiaodong and Wang, Xiaohan and Shen, Xiaojin and Chen, Xiaokang and Zhang, Xiaokang and Chen, Xiaosha and Nie, Xiaotao and Sun, Xiaowen and Wang, Xiaoxiang and Cheng, Xin and Liu, Xin and Xie, Xin and Liu, Xingchao and Yu, Xingkai and Song, Xinnan and Shan, Xinxia and Zhou, Xinyi and Yang, Xinyu and Li, Xinyuan and Su, Xuecheng and Lin, Xuheng and Li, Y. K. and Wang, Y. Q. and Wei, Y. X. and Zhu, Y. X. and Zhang, Yang and Xu, Yanhong and Xu, Yanhong and Huang, Yanping and Li, Yao and Zhao, Yao and Sun, Yaofeng and Li, Yaohui and Wang, Yaohui and Yu, Yi and Zheng, Yi and Zhang, Yichao and Shi, Yifan and Xiong, Yiliang and He, Ying and Tang, Ying and Piao, Yishi and Wang, Yisong and Tan, Yixuan and Ma, Yiyang and Liu, Yiyuan and Guo, Yongqiang and Wu, Yu and Ou, Yuan and Zhu, Yuchen and Wang, Yuduan and Gong, Yue and Zou, Yuheng and He, Yujia and Zha, Yukun and Xiong, Yunfan and Ma, Yunxian and Yan, Yuting and Luo, Yuxiang and You, Yuxiang and Liu, Yuxuan and Zhou, Yuyang and Wu, Z. F. and Ren, Z. Z. and Ren, Zehui and Sha, Zhangli and Fu, Zhe and Xu, Zhean and Huang, Zhen and Zhang, Zhen and Xie, Zhenda and Zhang, Zhengyan and Hao, Zhewen and Gou, Zhibin and Ma, Zhicheng and Yan, Zhigang and Shao, Zhihong and Xu, Zhipeng and Wu, Zhiyu and Zhang, Zhongyu and Li, Zhuoshu and Gu, Zihui and Zhu, Zijia and Liu, Zijun and Li, Zilin and Xie, Ziwei and Song, Ziyang and Gao, Ziyi and Pan, Zizheng},
  year = 2025,
  month = feb,
  number = {arXiv:2412.19437},
  eprint = {2412.19437},
  primaryclass = {cs},
  publisher = {arXiv},
  doi = {10.48550/arXiv.2412.19437},
  urldate = {2026-01-21},
  archiveprefix = {arXiv},
}

@misc{cheng2025mone,
      title={Mixture of Neuron Experts}, 
      author={Runxi Cheng and Yuchen Guan and Yucheng Ding and Qingguo Hu and Yongxian Wei and Chun Yuan and Yelong Shen and Weizhu Chen and Yeyun Gong},
      year={2025},
      eprint={2510.05781},
      archivePrefix={arXiv},
      primaryClass={cs.CL},
      url={https://arxiv.org/abs/2510.05781}, 
}

@article{fedus22switch,
  author       = {William Fedus and
                  Barret Zoph and
                  Noam Shazeer},
  title        = {Switch Transformers: Scaling to Trillion Parameter Models with Simple
                  and Efficient Sparsity},
  journal      = {J. Mach. Learn. Res.},
  volume       = {23},
  pages        = {120:1--120:39},
  year         = {2022},
  url          = {https://jmlr.org/papers/v23/21-0998.html},
  bibsource    = {dblp computer science bibliography, https://dblp.org}
}

@inproceedings{shazeer17outrageous,
  author       = {Noam Shazeer and
                  Azalia Mirhoseini and
                  Krzysztof Maziarz and
                  Andy Davis and
                  Quoc V. Le and
                  Geoffrey E. Hinton and
                  Jeff Dean},
  title        = {Outrageously Large Neural Networks: The Sparsely-Gated Mixture-of-Experts
                  Layer},
  booktitle    = {5th International Conference on Learning Representations, {ICLR} 2017,
                  Toulon, France, April 24-26, 2017, Conference Track Proceedings},
  publisher    = {OpenReview.net},
  year         = {2017},
  url          = {https://openreview.net/forum?id=B1ckMDqlg},
  bibsource    = {dblp computer science bibliography, https://dblp.org}
}

@article{he2024peer,
  author       = {Xu Owen He},
  title        = {Mixture of {A} Million Experts},
  journal      = {CoRR},
  volume       = {abs/2407.04153},
  year         = {2024},
  url          = {https://doi.org/10.48550/arXiv.2407.04153},
  doi          = {10.48550/ARXIV.2407.04153},
  eprinttype    = {arXiv},
  eprint       = {2407.04153},
  bibsource    = {dblp computer science bibliography, https://dblp.org}
}

@inproceedings{nogueira2024mowe,
    title = "Memory Augmented Language Models through Mixture of Word Experts",
    author = "Nogueira dos Santos, Cicero  and
      Lee-Thorp, James  and
      Noble, Isaac  and
      Chang, Chung-Ching  and
      Uthus, David",
    editor = "Duh, Kevin  and
      Gomez, Helena  and
      Bethard, Steven",
    booktitle = "Proceedings of the 2024 Conference of the North American Chapter of the Association for Computational Linguistics: Human Language Technologies (Volume 1: Long Papers)",
    month = jun,
    year = "2024",
    address = "Mexico City, Mexico",
    publisher = "Association for Computational Linguistics",
    url = "https://aclanthology.org/2024.naacl-long.249/",
    doi = "10.18653/v1/2024.naacl-long.249",
    pages = "4425--4438",
}

@misc{nguyen2025ondeepseekmoe,
  title = {On {{DeepSeekMoE}}: {{Statistical Benefits}} of {{Shared Experts}} and {{Normalized Sigmoid Gating}}},
  shorttitle = {On {{DeepSeekMoE}}},
  author = {Nguyen, Huy and Doan, Thong T. and Pham, Quang and Bui, Nghi D. Q. and Ho, Nhat and Rinaldo, Alessandro},
  year = 2025,
  month = may,
  number = {arXiv:2505.10860},
  eprint = {2505.10860},
  primaryclass = {cs},
  publisher = {arXiv},
  doi = {10.48550/arXiv.2505.10860},
  urldate = {2026-01-20},
  archiveprefix = {arXiv},
}

@misc{rajbhandari2022deepspeed,
  title = {{{DeepSpeed-MoE}}: {{Advancing Mixture-of-Experts Inference}} and {{Training}} to {{Power Next-Generation AI Scale}}},
  shorttitle = {{{DeepSpeed-MoE}}},
  author = {Rajbhandari, Samyam and Li, Conglong and Yao, Zhewei and Zhang, Minjia and Aminabadi, Reza Yazdani and Awan, Ammar Ahmad and Rasley, Jeff and He, Yuxiong},
  year = 2022,
  month = jul,
  number = {arXiv:2201.05596},
  eprint = {2201.05596},
  primaryclass = {cs},
  publisher = {arXiv},
  doi = {10.48550/arXiv.2201.05596},
  urldate = {2026-01-20},
  archiveprefix = {arXiv},
}

@inproceedings{ainslie2023gqa,
  author       = {Joshua Ainslie and
                  James Lee{-}Thorp and
                  Michiel de Jong and
                  Yury Zemlyanskiy and
                  Federico Lebr{\'{o}}n and
                  Sumit Sanghai},
  editor       = {Houda Bouamor and
                  Juan Pino and
                  Kalika Bali},
  title        = {{GQA:} Training Generalized Multi-Query Transformer Models from Multi-Head
                  Checkpoints},
  booktitle    = {Proceedings of the 2023 Conference on Empirical Methods in Natural
                  Language Processing, {EMNLP} 2023, Singapore, December 6-10, 2023},
  pages        = {4895--4901},
  publisher    = {Association for Computational Linguistics},
  year         = {2023},
  url          = {https://doi.org/10.18653/v1/2023.emnlp-main.298},
  doi          = {10.18653/V1/2023.EMNLP-MAIN.298},
  bibsource    = {dblp computer science bibliography, https://dblp.org}
}

@misc{he2021fastmoe,
  title = {{{FastMoE}}: {{A Fast Mixture-of-Expert Training System}}},
  shorttitle = {{{FastMoE}}},
  author = {He, Jiaao and Qiu, Jiezhong and Zeng, Aohan and Yang, Zhilin and Zhai, Jidong and Tang, Jie},
  year = 2021,
  month = mar,
  number = {arXiv:2103.13262},
  eprint = {2103.13262},
  primaryclass = {cs},
  publisher = {arXiv},
  doi = {10.48550/arXiv.2103.13262},
  archiveprefix = {arXiv}
}

@misc{tan2024scattermoe,
  title = {Scattered {{Mixture-of-Experts Implementation}}},
  author = {Tan, Shawn and Shen, Yikang and Panda, Rameswar and Courville, Aaron},
  year = 2024,
  month = oct,
  number = {arXiv:2403.08245},
  eprint = {2403.08245},
  primaryclass = {cs},
  publisher = {arXiv},
  doi = {10.48550/arXiv.2403.08245},
  archiveprefix = {arXiv}
}

@inproceedings{zheng2023pit,
  author       = {Ningxin Zheng and
                  Huiqiang Jiang and
                  Quanlu Zhang and
                  Zhenhua Han and
                  Lingxiao Ma and
                  Yuqing Yang and
                  Fan Yang and
                  Chengruidong Zhang and
                  Lili Qiu and
                  Mao Yang and
                  Lidong Zhou},
  editor       = {Jason Flinn and
                  Margo I. Seltzer and
                  Peter Druschel and
                  Antoine Kaufmann and
                  Jonathan Mace},
  title        = {{PIT:} Optimization of Dynamic Sparse Deep Learning Models via Permutation
                  Invariant Transformation},
  booktitle    = {Proceedings of the 29th Symposium on Operating Systems Principles,
                  {SOSP} 2023, Koblenz, Germany, October 23-26, 2023},
  pages        = {331--347},
  publisher    = {{ACM}},
  year         = {2023},
  url          = {https://doi.org/10.1145/3600006.3613139},
  doi          = {10.1145/3600006.3613139},
  bibsource    = {dblp computer science bibliography, https://dblp.org}
}

@misc{liu2025surveyinferenceopt,
  title = {A {{Survey}} on {{Inference Optimization Techniques}} for {{Mixture}} of {{Experts Models}}},
  author = {Liu, Jiacheng and Tang, Peng and Wang, Wenfeng and Ren, Yuhang and Hou, Xiaofeng and Heng, Pheng-Ann and Guo, Minyi and Li, Chao},
  year = 2025,
  month = jan,
  number = {arXiv:2412.14219},
  eprint = {2412.14219},
  primaryclass = {cs},
  publisher = {arXiv},
  doi = {10.48550/arXiv.2412.14219},
  archiveprefix = {arXiv},
  langid = {english}
}

@misc{mu2025surveymoe,
  title = {A {{Comprehensive Survey}} of {{Mixture-of-Experts}}: {{Algorithms}}, {{Theory}}, and {{Applications}}},
  shorttitle = {A {{Comprehensive Survey}} of {{Mixture-of-Experts}}},
  author = {Mu, Siyuan and Lin, Sen},
  year = 2025,
  month = apr,
  number = {arXiv:2503.07137},
  eprint = {2503.07137},
  primaryclass = {cs},
  publisher = {arXiv},
  doi = {10.48550/arXiv.2503.07137},
  archiveprefix = {arXiv},
  langid = {english}
}

@misc{sun2025speedsurvey,
  title = {Speed {{Always Wins}}: {{A Survey}} on {{Efficient Architectures}} for {{Large Language Models}}},
  shorttitle = {Speed {{Always Wins}}},
  author = {Sun, Weigao and Hu, Jiaxi and Zhou, Yucheng and Du, Jusen and Lan, Disen and Wang, Kexin and Zhu, Tong and Qu, Xiaoye and Zhang, Yu and Mo, Xiaoyu and Liu, Daizong and Liang, Yuxuan and Chen, Wenliang and Li, Guoqi and Cheng, Yu},
  year = 2025,
  month = aug,
  number = {arXiv:2508.09834},
  eprint = {2508.09834},
  primaryclass = {cs},
  publisher = {arXiv},
  doi = {10.48550/arXiv.2508.09834},
  archiveprefix = {arXiv}
}

@inproceedings{gale2023meagblocks,
  author       = {Trevor Gale and
                  Deepak Narayanan and
                  Cliff Young and
                  Matei Zaharia},
  editor       = {Dawn Song and
                  Michael Carbin and
                  Tianqi Chen},
  title        = {MegaBlocks: Efficient Sparse Training with Mixture-of-Experts},
  booktitle    = {Proceedings of the Sixth Conference on Machine Learning and Systems,
                  MLSys 2023, Miami, FL, USA, June 4-8, 2023},
  publisher    = {mlsys.org},
  year         = {2023},
  bibsource    = {dblp computer science bibliography, https://dblp.org}
}

@inproceedings{li2023lazyneuron,
  author       = {Zonglin Li and
                  Chong You and
                  Srinadh Bhojanapalli and
                  Daliang Li and
                  Ankit Singh Rawat and
                  Sashank J. Reddi and
                  Ke Ye and
                  Felix Chern and
                  Felix X. Yu and
                  Ruiqi Guo and
                  Sanjiv Kumar},
  title        = {The Lazy Neuron Phenomenon: On Emergence of Activation Sparsity in
                  Transformers},
  booktitle    = {The Eleventh International Conference on Learning Representations,
                  {ICLR} 2023, Kigali, Rwanda, May 1-5, 2023},
  publisher    = {OpenReview.net},
  year         = {2023},
  url          = {https://openreview.net/forum?id=TJ2nxciYCk-},
  bibsource    = {dblp computer science bibliography, https://dblp.org}
}

@article{zhou2025dern,
  author       = {Yixiao Zhou and
                  Ziyu Zhao and
                  Dongzhou Cheng and
                  Zhiliang Wu and
                  Jie Gui and
                  Yi Yang and
                  Fei Wu and
                  Yu Cheng and
                  Hehe Fan},
  title        = {Dropping Experts, Recombining Neurons: Retraining-Free Pruning for
                  Sparse Mixture-of-Experts LLMs},
  journal      = {CoRR},
  volume       = {abs/2509.10377},
  year         = {2025},
  url          = {https://doi.org/10.48550/arXiv.2509.10377},
  doi          = {10.48550/ARXIV.2509.10377},
  eprinttype    = {arXiv},
  eprint       = {2509.10377},
  bibsource    = {dblp computer science bibliography, https://dblp.org}
}

@inproceedings{yang2024xmoe,
    title = "{XM}o{E}: Sparse Models with Fine-grained and Adaptive Expert Selection",
    author = "Yang, Yuanhang  and
      Qi, Shiyi  and
      Gu, Wenchao  and
      Wang, Chaozheng  and
      Gao, Cuiyun  and
      Xu, Zenglin",
    editor = "Ku, Lun-Wei  and
      Martins, Andre  and
      Srikumar, Vivek",
    booktitle = "Findings of the Association for Computational Linguistics: ACL 2024",
    month = aug,
    year = "2024",
    address = "Bangkok, Thailand",
    publisher = "Association for Computational Linguistics",
    url = "https://aclanthology.org/2024.findings-acl.694/",
    doi = "10.18653/v1/2024.findings-acl.694",
    pages = "11664--11674",
}

@misc{guo2025sonicmoe,
      title={SonicMoE: Accelerating MoE with IO and Tile-aware Optimizations}, 
      author={Wentao Guo and Mayank Mishra and Xinle Cheng and Ion Stoica and Tri Dao},
      year={2025},
      eprint={2512.14080},
      archivePrefix={arXiv},
      primaryClass={cs.LG},
      url={https://arxiv.org/abs/2512.14080},
}

@misc{cobbe2021trainingverifierssolvemath,
      title={Training Verifiers to Solve Math Word Problems}, 
      author={Karl Cobbe and Vineet Kosaraju and Mohammad Bavarian and Mark Chen and Heewoo Jun and Lukasz Kaiser and Matthias Plappert and Jerry Tworek and Jacob Hilton and Reiichiro Nakano and Christopher Hesse and John Schulman},
      year={2021},
      eprint={2110.14168},
      archivePrefix={arXiv},
      primaryClass={cs.LG},
      url={https://arxiv.org/abs/2110.14168}, 
}

@misc{hendrycks2021measuringmathematicalproblemsolving,
      title={Measuring Mathematical Problem Solving With the MATH Dataset}, 
      author={Dan Hendrycks and Collin Burns and Saurav Kadavath and Akul Arora and Steven Basart and Eric Tang and Dawn Song and Jacob Steinhardt},
      year={2021},
      eprint={2103.03874},
      archivePrefix={arXiv},
      primaryClass={cs.LG},
      url={https://arxiv.org/abs/2103.03874}, 
}

@misc{austin2021programsynthesislargelanguage,
      title={Program Synthesis with Large Language Models}, 
      author={Jacob Austin and Augustus Odena and Maxwell Nye and Maarten Bosma and Henryk Michalewski and David Dohan and Ellen Jiang and Carrie Cai and Michael Terry and Quoc Le and Charles Sutton},
      year={2021},
      eprint={2108.07732},
      archivePrefix={arXiv},
      primaryClass={cs.PL},
      url={https://arxiv.org/abs/2108.07732}, 
}

@misc{chen2021codex,
  title={Evaluating Large Language Models Trained on Code},
  author={Mark Chen and Jerry Tworek and Heewoo Jun and Qiming Yuan and Henrique Ponde de Oliveira Pinto and Jared Kaplan and Harri Edwards and Yuri Burda and Nicholas Joseph and Greg Brockman and Alex Ray and Raul Puri and Gretchen Krueger and Michael Petrov and Heidy Khlaaf and Girish Sastry and Pamela Mishkin and Brooke Chan and Scott Gray and Nick Ryder and Mikhail Pavlov and Alethea Power and Lukasz Kaiser and Mohammad Bavarian and Clemens Winter and Philippe Tillet and Felipe Petroski Such and Dave Cummings and Matthias Plappert and Fotios Chantzis and Elizabeth Barnes and Ariel Herbert-Voss and William Hebgen Guss and Alex Nichol and Alex Paino and Nikolas Tezak and Jie Tang and Igor Babuschkin and Suchir Balaji and Shantanu Jain and William Saunders and Christopher Hesse and Andrew N. Carr and Jan Leike and Josh Achiam and Vedant Misra and Evan Morikawa and Alec Radford and Matthew Knight and Miles Brundage and Mira Murati and Katie Mayer and Peter Welinder and Bob McGrew and Dario Amodei and Sam McCandlish and Ilya Sutskever and Wojciech Zaremba},
  year={2021},
  eprint={2107.03374},
  archivePrefix={arXiv},
  primaryClass={cs.LG}
}

@misc{fang2025klotskiefficientmixtureofexpertinference,
      title={Klotski: Efficient Mixture-of-Expert Inference via Expert-Aware Multi-Batch Pipeline}, 
      author={Zhiyuan Fang and Yuegui Huang and Zicong Hong and Yufeng Lyu and Wuhui Chen and Yue Yu and Fan Yu and Zibin Zheng},
      year={2025},
      eprint={2502.06888},
      archivePrefix={arXiv},
      primaryClass={cs.LG},
      url={https://arxiv.org/abs/2502.06888}, 
}

@misc{he2026expertflowefficientmixtureofexpertsinference,
      title={ExpertFlow: Efficient Mixture-of-Experts Inference via Predictive Expert Caching and Token Scheduling}, 
      author={Xin He and Shunkang Zhang and Kaijie Tang and Shaohuai Shi and Yuxin Wang and Zihao Zeng and Zhenheng Tang and Xiaowen Chu and Haiyan Yin and Ivor W. Tsang and Yew Soon Ong},
      year={2026},
      eprint={2410.17954},
      archivePrefix={arXiv},
      primaryClass={cs.AI},
      doi={https://doi.org/10.1145/3770743.3804292},
      url={https://arxiv.org/abs/2410.17954}, 
}
\bibliographystyle{icml2026_etal}

\newpage
\appendix
\setcounter{table}{0}
\renewcommand{\thetable}{\Alph{table}}
\setcounter{figure}{0}
\renewcommand{\thefigure}{\Alph{figure}}
\setcounter{equation}{0}
\renewcommand{\theequation}{\roman{equation}}
\onecolumn

\section{Complexity Analysis}
\label{app:complexity_analysis}

In this section, we provide a detailed complexity derivation for both the Cartesian Product Router and the Expert-Centric Scheduling strategy. We focus on theoretical FLOPs for routing and Memory Traffic (I/O) for scheduling, as these correspond to the primary bottlenecks in each stage.

\paragraph{Routing Complexity Derivation.}
Consider a standard Top-$K$ router with $N$ experts and hidden dimension $d$. For a single token $x \in \mathbb{R}^d$, the router computes logits $h = x W_g$, where $W_g \in \mathbb{R}^{d \times N}$.
The computational complexity (FLOPs) for the projection is:
\begin{equation}
\mathcal{C}_{\text{std}} = 2 \cdot d \cdot N
\end{equation}
The parameter storage requirement is $\mathcal{M}_{\text{std}} = d \cdot N$.
With $N$ scaling to millions (e.g., $10^6$), both storage and computation become prohibitive (e.g., $2 \cdot 10^9$ FLOPs per token just for routing).

The proposed \textbf{Cartesian Product Router} decomposes the expert index space into a grid of size $N_r \times N_c$ (where $N = N_r \cdot N_c$). It employs two projection matrices $W_r \in \mathbb{R}^{d \times N_r}$ and $W_c \in \mathbb{R}^{d \times N_c}$.
The logit computation involves two smaller projections: $s_r = x W_r$ and $s_c = x W_c$.
The computational complexity becomes:
\begin{equation}
\mathcal{C}_{\text{cart}} = 2 \cdot d \cdot (N_r + N_c)
\end{equation}
Assuming a balanced grid where $N_r \approx N_c \approx \sqrt{N}$, the complexity is:
\begin{equation}
\mathcal{C}_{\text{cart}} \approx 4 \cdot d \cdot \sqrt{N}
\end{equation}
Comparing the two, the reduction factor is:
\begin{equation}
\frac{\mathcal{C}_{\text{std}}}{\mathcal{C}_{\text{cart}}} = \frac{2dN}{4d\sqrt{N}} = \frac{\sqrt{N}}{2}
\end{equation}
For $N=10^6$, this yields a theoretical speedup of $500\times$ in router projection FLOPs. Similarly, the parameter storage scales with $O(\sqrt{N}d)$ instead of $O(Nd)$, making million-scale expert routing feasible.

\paragraph{Top-$K$ Selection Complexity.}
While the Cartesian Product Router efficiently reduces projection complexity, selecting the top-$K$ experts from $N$ scores remains a challenge. Standard approaches materialize the full score matrix and sort it globally, costing $O(N)$ memory traffic and $O(N \log N)$ or $O(N)$ operations.
Our implementation utilizes a fused \textbf{Block-wise Merge Selection} kernel that avoids full score materialization.
We partition the $N$ experts into blocks of size $B_{sel}$ (e.g., 4096).
For each block, the kernel:
(1) Computes scores on-the-fly (Complexity: $O(B_{sel})$).
(2) Performs iterative max-reduction to find the local top-$K$ candidates (Complexity: $O(B_{sel} \cdot K)$).
(3) Merges local candidates with the global top-$K$ buffer (Complexity: $O(K^2)$).
Summing over $N/B_{sel}$ blocks, the total time complexity per token is:
\begin{equation}
\mathcal{C}_{\text{select}} = \frac{N}{B_{sel}} \cdot (B_{sel} \cdot K + K^2) = O(N \cdot K + \frac{N}{B_{sel}} K^2)
\end{equation}
The $O(K^2)$ term arises from merging local candidates into the global buffer, typically implemented via insertion sort within GPU registers.
Our fused approach eliminates the dominant $O(N)$ global memory I/O bottleneck. By keeping scores in registers/SRAM, the operation becomes compute-bound and effectively negligible in latency on modern GPUs.

\paragraph{Memory Traffic Analysis for Scheduling.}
We analyze the memory I/O volume for the routed FFN execution. Let $L$ be the number of tokens in a batch, and $K$ be the number of activated experts per token.
The computation involves retrieving expert parameters $W, V \in \mathbb{R}^{N \times d}$ and performing the forward pass.
We assume the worst-case scenario for baselines where no cache reuse occurs due to large $N$ and scattered access patterns.

\noindent\textbf{Token-Centric Scheduling (Baseline).}
In this paradigm, each token $x_l$ retrieves parameters for its specific top-$K$ selected experts indices $\mathcal{I}_l = \{I_{l,0}, \dots, I_{l,K-1}\}$.
The total memory traffic for loading expert parameters is proportional to the total number of expert executions:
\begin{equation}
\mathcal{D}_{\text{token}} \propto \sum_{l=0}^{L-1} \sum_{k=0}^{K-1} (Size(W_{I_{l,k}}) + Size(V_{I_{l,k}})) = 2 d \cdot L \cdot K
\end{equation}
This approach suffers from redundancy when multiple tokens select the same expert. Furthermore, the memory accesses are non-contiguous (gather operations), significantly degrading effective bandwidth utilization.

\noindent\textbf{Expert-Centric Scheduling (Ours).}
This strategy inverts the loop order. We first identify the set of unique activated experts in the batch: $\mathbb{E}_{\text{active}} = \bigcup_{l=0}^{L-1} \mathcal{I}_l$.
The execution groups all tokens assigned to a specific expert $E_j \in \mathbb{E}_{\text{active}}$.
Consequently, the parameters for expert $E_j$ are loaded exactly once from HBM to SRAM.
The total memory traffic for expert parameters becomes:
\begin{equation}
\mathcal{D}_{\text{expert}} \propto \sum_{j \in \mathbb{E}_{\text{active}}} (Size(W_{j}) + Size(V_{j})) = 2 d \cdot |\mathbb{E}_{\text{active}}|
\end{equation}
The reduction in memory traffic is defined by the ratio:
\begin{equation}
\eta = \frac{\mathcal{D}_{\text{token}}}{\mathcal{D}_{\text{expert}}} = \frac{L \cdot K}{|\mathbb{E}_{\text{active}}|}
\end{equation}
In our \method settings, where experts are fine-grained but $K$ is large, each expert is frequently accessed by multiple tokens in a batch (i.e., $L \cdot K \gg |\mathbb{E}_{\text{active}}|$). Thus $\eta \gg 1$, indicating a substantial reduction in parameter I/O.

\paragraph{Scheduling Overhead and Token Traffic.}
While expert-centric scheduling optimizes parameter loading, it introduces a reordering step and necessitates token reloading. We analyze these costs below:

\textit{1. Sort Overhead:}
The preprocessing involves flattening the routing indices and sorting $M = L \times K$ tasks by (Expert Group, Token ID). The complexity is $O(M \log M)$. Given that GPU memory bandwidth is the primary bottleneck, this lightweight integer sorting (performed efficiently via radix sort) is negligible. Empirically, scheduling occupies $<5\%$ of total latency, well-amortized by the speedup in the GEMM phase.

\textit{2. Token Memory Access:}
A potential concern is that if a token activates experts across multiple groups, it must be loaded multiple times. However, our hierarchical sorting ensures that within each expert group, tokens are processed in \emph{increasing order of Token ID}. This converts token access into a \textbf{strictly sequential stream}, enabling perfectly coalesced memory reads. Unlike the random access patterns in token-centric baselines, our approach fully utilizes the high sequential bandwidth of HBM. Furthermore, since $K$ is large, tasks for the same token often cluster in consecutive groups, allowing effectively cached reuse.

\section{Experimental Setup}
\label{app:explicit_setup}

\begin{table*}[b]
  \centering
  \caption{
    \textbf{Speed and Memory Benchmarking Configurations}.
    We list the key hyperparameters used for measuring inference latency and memory usage.
    Total Params denotes the total parameter count of the FFN layer, while Act Params refers to the number of parameters active during the forward pass of a single token.
    Act Tokens represents the number of tokens in a batch.
    $d$ is the hidden dimension size.
    $N$ is the total number of experts.
    $d_{rffn}$ and $d_{sffn}$ denote the intermediate dimensions of the routed FFN and shared FFN (if applicable), respectively.
    $K$ indicates the number of experts selected per token.
  }
  \resizebox{\linewidth}{!}
  {
    \begin{tabular}{@{}ccccccccccccc@{}}
    \toprule
    \sc{Algo} & \sc{Total Params} & \sc{Act Params} & \sc{Act Tokens} & $d$ & $N$ & $d_{rffn}$ & $d_{sffn}$ & $K$ \\
    \midrule
    Gshard & ~200M & [3M, 6M, 12M, 25M] & [1k, 2k, 4k, 8k, 16k] & 1024 & 128 & 512 & - & [2, 4, 8, 16] \\
    DeepSeekMoE & ~200M & [6M, 9M, 15M, 28M] & [1k, 2k, 4k, 8k, 16k] & 1024 & 256 & 256 & 1024 & [4, 8, 16, 32] \\
    PKM & ~200M & [3M, 5M, 13M, 26M] & [1k, 2k, 4k, 8k, 16k] & 1024 & 197136 & - & - & [2500, 5000, 12500, 25000] \\
    PEER & ~200M & [3M, 5M, 13M, 26M] & [1k, 2k, 4k, 8k, 16k] & 1024 & 102400 & - & - & [1250, 2500, 6250, 12500] \\
    \method & ~200M & [4M, 7M, 15M, 28M] & [1k, 2k, 4k, 8k, 16k] & 1024 & 102400 & - & 1024 & [512, 1024, 2048, 4096] \\
    \bottomrule
    \end{tabular}
  }
  \label{table:speed_memory_configs}
\end{table*}

\begin{table*}[!t]
  \centering
  \caption{
    \textbf{Language Modeling Configurations}.
    Detailed hyperparameters for pre-training experiments across different scales (Activation 80M, 200M, 680M, and 1.7B).
    Total Params and Act Params indicate the total model size and the per-token active parameter count, respectively.
    Number of Step and Number of Batch specify the training duration and total tokens seen.
    LR is the peak learning rate.
    $n_{layer}$ and $d_{model}$ denote the number of transformer layers and the hidden dimension.
    Tied Emb indicates whether the input and output embeddings are tied.
  }
  \resizebox{\linewidth}{!}
  {
    \begin{tabular}{@{}ccccccccc@{}}
    \toprule
    \sc{Algo} & \sc{Total Params} & \sc{Act Params} & \sc{Number of Step} & \sc{Batch Tokens} & \sc{LR} & $n_{layer}$ & $d_{model}$ & \sc{Tied Emb} \\
    \midrule
    \multicolumn{9}{c}{\text{Activation 80M}} \\
    \midrule
    Dense & ~80M & ~80M & 13500 & 0.128M tokens & 3e-3 & 12 & 768 & \cmark \\
    Gshard & ~280M & ~80M & 13500 & 0.128M tokens & 3e-3 & 12 & 768 & \cmark \\
    DeepSeekMoE & ~280M & ~80M & 13500 & 0.128M tokens & 3e-3 & 12 & 768 & \cmark \\
    PKM & ~280M & ~80M & 13500 & 0.128M tokens & 3e-3 & 12 & 768 & \cmark \\
    PEER & ~280M & ~80M & 13500 & 0.128M tokens & 3e-3 & 12 & 768 & \cmark \\
    \method & ~280M & ~80M & 13500 & 0.128M tokens & 3e-3 & 12 & 768 & \cmark \\
    \midrule
    \multicolumn{9}{c}{\text{Activation 200M}} \\
    \midrule
    Dense & ~200M & ~200M & 20800 & 0.192M tokens & 2e-3 & 16 & 1024 & \cmark \\
    Gshard & ~800M & ~200M & 20800 & 0.192M tokens & 2e-3 & 16 & 1024 & \cmark \\
    DeepSeekMoE & ~800M & ~200M & 20800 & 0.192M tokens & 2e-3 & 16 & 1024 & \cmark \\
    PKM & ~800M & ~200M & 20800 & 0.192M tokens & 2e-3 & 16 & 1024 & \cmark \\
    PEER & ~800M & ~200M & 20800 & 0.192M tokens & 2e-3 & 16 & 1024 & \cmark \\
    \method & ~800M & ~200M & 20800 & 0.192M tokens & 2e-3 & 16 & 1024 & \cmark \\
    \midrule
    \multicolumn{9}{c}{\text{Activation 680M}} \\
    \midrule
    Dense & ~680M & ~680M & 35000 & 0.392M tokens & 1e-3 & 24 & 1536 & \cmark \\
    Gshard & ~2.7B & ~680M & 35000 & 0.392M tokens & 1e-3 & 24 & 1536 & \cmark \\
    DeepSeekMoE & ~2.7B & ~680M & 35000 & 0.392M tokens & 1e-3 & 24 & 1536 & \cmark \\
    PKM & ~2.7B & ~680M & 35000 & 0.392M tokens & 1e-3 & 24 & 1536 & \cmark \\
    PEER & ~2.7B & ~680M & 35000 & 0.392M tokens & 1e-3 & 24 & 1536 & \cmark \\
    \method & ~2.7B & ~680M & 35000 & 0.392M tokens & 1e-3 & 24 & 1536 & \cmark \\
    \midrule
    \multicolumn{9}{c}{\text{Activation 1.7B}} \\
    \midrule 
    Dense & ~1.7B & ~1.7B & 40000 & 1M tokens & 1e-3 & 28 & 2048 & \xmark \\
    Gshard & ~6.4B & ~1.7B & 40000 & 1M tokens & 1e-3 & 28 & 2048 & \xmark \\
    DeepSeekMoE & ~6.4B & ~1.7B & 40000 & 1M tokens & 1e-3 & 28 & 2048 & \xmark \\
    PKM & ~6.4B & ~1.7B & 40000 & 1M tokens & 1e-3 & 28 & 2048 & \xmark \\
    PEER & ~6.4B & ~1.7B & 40000 & 1M tokens & 1e-3 & 28 & 2048 & \xmark \\
    \method & ~6.4B & ~1.7B & 40000 & 1M tokens & 1e-3 & 28 & 2048 & \xmark \\
    \bottomrule
    \end{tabular}
  }
  \label{table:lm_configs}
\end{table*}

In this section, we provide detailed configurations for the experiments conducted in Sec.~\ref{sec:experiments}.
To ensure a rigorous evaluation, we prioritize \textbf{architectural comparison} via controlled pre-training from scratch. Rather than comparing against off-the-shelf checkpoints, which differ in training data and recipes, we represent baselines using their underlying architectural prototypes (e.g., Gshard, DeepSeekMoE) trained on a strictly identical corpus. This isolates the impact of the MoE architectural design from confounding factors.\label{app:baseline_justification} Although academic resource constraints limit pre-training to the 1.7B activated-parameter scale, we verify that all methods adhere to predictable scaling laws (Section~\ref{subsec:main_results}), ensuring our findings extrapolate to larger scales.

Detailed implementation specifications for all evaluated methods are as follows: For coarse-grained baselines (Gshard, DeepSeekMoE), we adopt the best-performance kernels released by NVIDIA in CuTile. For fine-grained baselines (PKM, PEER), we employ highly-optimized Triton fused kernels. For \method, we implement our custom Expert-Centric Scheduling using Triton to maximize hardware utilization.

Table~\ref{table:speed_memory_configs} outlines the configurations used for \textbf{Speed and Memory Benchmarking}. 
In this benchmark, we evaluate system performance by varying one dimension while keeping the other fixed at its minimum value. 
Specifically, when sweeping the activated parameter budget (Act Params), the number of activated tokens (Act Tokens) is fixed at 1K. 
Conversely, when varying the number of activated tokens, the activated parameter budget is held at the minimum configuration for each method.
Table~\ref{table:lm_configs} presents the comprehensive \textbf{Language Modeling Configurations} for different model scales. 
We detail the model architecture hyperparameters (total parameters, activated parameters, layer count, model dimension, etc.) and training recipes (learning rate, batch size, training steps) to ensure reproducibility.

\section{Expert Parallelism Communication Overhead}
\label{app:ep_communication}

To verify the communication efficiency of \method in large-scale distributed training scenarios, we conducted distributed experiments focused on the Expert Parallelism (EP) communication overhead. 
We investigated the impact of the number of experts and sequence length on communication bandwidth. 

\begin{figure}[t]
    \centering
    \includegraphics[width=\linewidth]{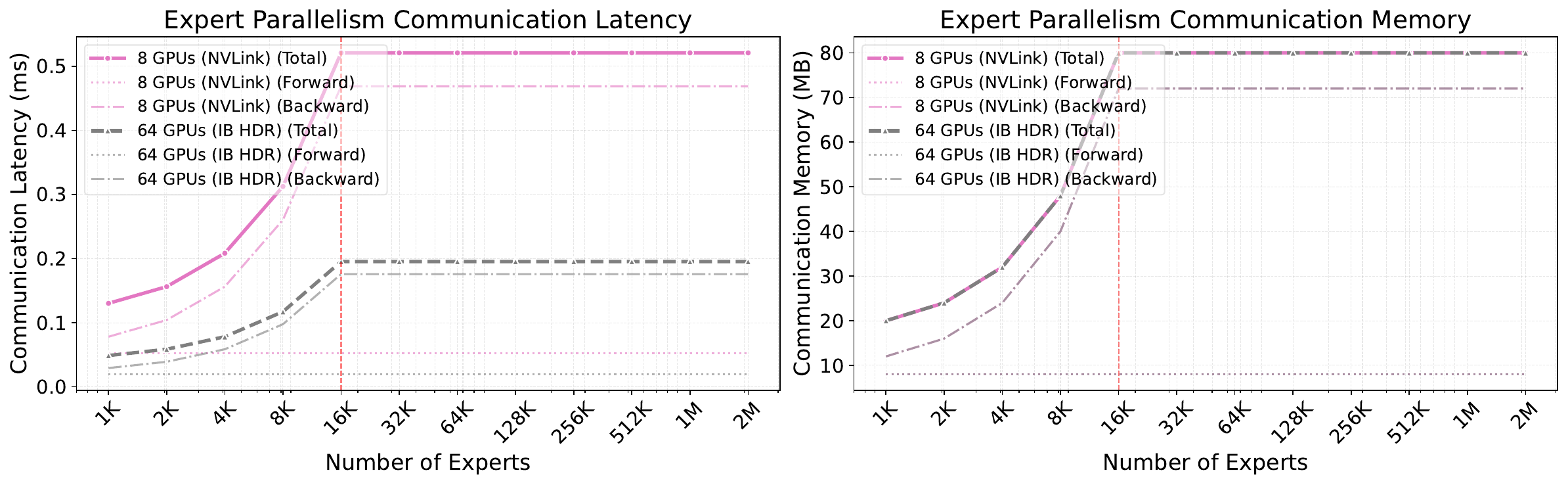}
    \caption{\textbf{Communication overhead vs. Number of Experts.} The communication latency and memory usage saturates and remains constant as the number of experts increases beyond the activation count.}
    \label{fig:ep_experts}
\end{figure}

\paragraph{Scalability with Number of Experts.}
Figure~\ref{fig:ep_experts} illustrates the change in communication overhead as the number of experts scales from 1K to 2M under a fixed sequence length.

\noindent\textbf{Key Finding: Decoupling Communication Cost from Model Capacity.}
The experimental results reveal a significant ``saturation effect'': when the number of experts $N$ exceeds the total number of activated experts ($n_{tokens} \times K = 16,384$), the total communication volume for the backward pass stabilizes at approximately 80MB and does not grow with increasing $N$. 
This implies that \method successfully breaks the bottleneck where communication overhead grows linearly with model capacity in traditional architectures. 
In an 8-GPU environment, this communication latency is merely 0.521 ms, which is negligible. 
This demonstrates \method's capability to support scaling to millions of experts with constant communication cost.

\paragraph{Scalability with Sequence Length.}

\begin{figure}[t]
    \centering
    \includegraphics[width=\linewidth]{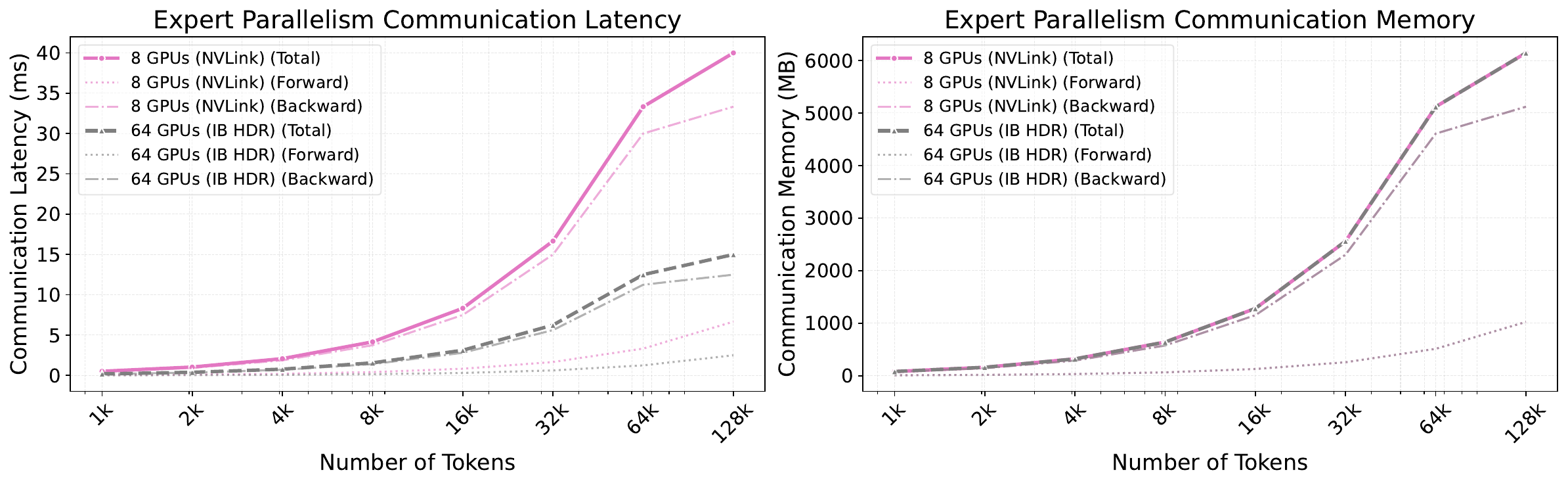}
    \caption{\textbf{Communication overhead vs. Sequence Length.} Communication volume scales linearly with sequence length.}
    \label{fig:ep_tokens}
\end{figure}

Figure~\ref{fig:ep_tokens} shows the communication overhead as the number of tokens scales from 1K to 128K under a fixed number of experts.

\noindent\textbf{Key Finding: Linear Communication Growth.}
Since the essence of EP communication is token distribution and gradient aggregation, the communication volume exhibits a strictly linear relationship with the sequence length $n_{tokens}$. 
Even in extreme scenarios with ultra-long sequences, where the total communication volume is approximately 6GB, the estimated latency on a 64-GPU cluster is only 15 ms. 
This indicates that \method's communication mechanism remains efficient for long-sequence training and does not become a bottleneck for training throughput.

\end{document}